\documentclass[10pt,journal,compsoc]{IEEEtran}
\ifCLASSOPTIONcompsoc
  \usepackage[nocompress]{cite}
\else
  \usepackage{cite}
\fi
\ifCLASSINFOpdf
\else
\fi
\usepackage{graphicx}

\usepackage{amsmath}
\usepackage{amssymb}
\usepackage{algorithmic}

\usepackage{array}
\usepackage{url}

\usepackage[colorlinks, citecolor=blue]{hyperref}

\usepackage{booktabs}
\usepackage{multirow}
\usepackage{makecell}

\newcommand{\etal}{\textit{et al.}}
\newcommand{\tablebodyfont}{\scriptsize}

\begin{document}
%
\title{CoGenRec: A Unified Generative Framework for Complementary Vision Reconstruction}

%
%
%
%

\author{Yapeng~Meng,
        Yuguo~Chen,
        Xiangru~Chen,
        Shaojun~Xu,
        Xuezhen~Wang,
        Lijian~Wang,
        Zheyu~Yang,
        Yihan~Lin,
        Luping~Shi,
        and~Rong~Zhao%
\thanks{Yapeng Meng, Yuguo Chen, Xiangru Chen, Shaojun Xu, Lijian Wang, Luping Shi, and Rong Zhao are with the Department of Precision Instrument, Tsinghua University, Beijing, China. Corresponding author: Rong Zhao (e-mail: r\_zhao@tsinghua.edu.cn), Yihan Lin (e-mail: linyh@xmu.edu.cn).}%
\thanks{Yihan Lin is with the Pen-Tung Sah Institute of Micro-Nano Science and Technology, Xiamen University, Xiamen, China.}%
\thanks{Xuezhen Wang is with The Hong Kong University of Science and Technology, HKSAR, China.}%
\thanks{Zheyu Yang is with Primevision Technology, Shanghai, China.}%
}

\IEEEtitleabstractindextext{%

\begin{abstract}
Capturing scenes with extreme motion and high dynamic range (HDR) is fundamentally constrained by trade-offs among temporal resolution, spatial detail, signal-to-noise ratio, and dynamic range under practical bandwidth, power, and quantum-efficiency limits. 
Conventional frame-based cameras therefore suffer from temporal undersampling, exposure-induced motion blur, and sensor saturation.
The Complementary Vision Sensor (CVS) addresses these limitations by decomposing visual information across two complementary pathways, a spatially dense, temporally sparse, long-exposure RGB pathway and a temporally dense, spatially sparse, short-exposure spatiotemporal difference pathway.
Their complementary measurements provide appearance, structural, and motion cues across a wide dynamic range that enable reconstruction beyond conventional frame-based imaging. 
We present CoGenRec, a complementarity-driven generative reconstruction framework that unifies video deblurring, high-temporal-resolution frame interpolation, and HDR fusion. CoGenRec encodes complementary measurements into multi-scale spatiotemporal features using exposure-aware appearance modeling and bidirectional temporal propagation. These features constrain a pre-trained video diffusion model and are injected into the VAE decoder to recover fine structures lost during latent-space compression, combining the scene fidelity of physical measurements with the appearance recovery capabilities of learned video priors.
To support generalization beyond simulated data, we construct a hybrid training dataset of approximately one million CVS clips by using a Digital Light Processing-based hardware conversion, noise-aware software simulation, and exposure augmentation.
On SportsSloMo-CVS, CoGenRec improves PSNR over the strongest general-purpose baseline for each task by 4.40 dB for deblurring, 3.89 dB for frame interpolation, and 5.04 dB for HDR fusion, respectively. It also achieves strong perceptual quality on 129 unseen scenes captured with real CVS hardware.
These results establish a unified approach to integrating complementary physical measurements with pre-trained generative video priors, advancing high-quality reconstruction under challenging motion and illumination conditions. Project page: \url{https://cogenrec.github.io/}.

\end{abstract}

}

\maketitle

\IEEEdisplaynontitleabstractindextext

%
\IEEEpeerreviewmaketitle

\begin{figure*}[t]
    \centering
    \includegraphics[width=\textwidth]{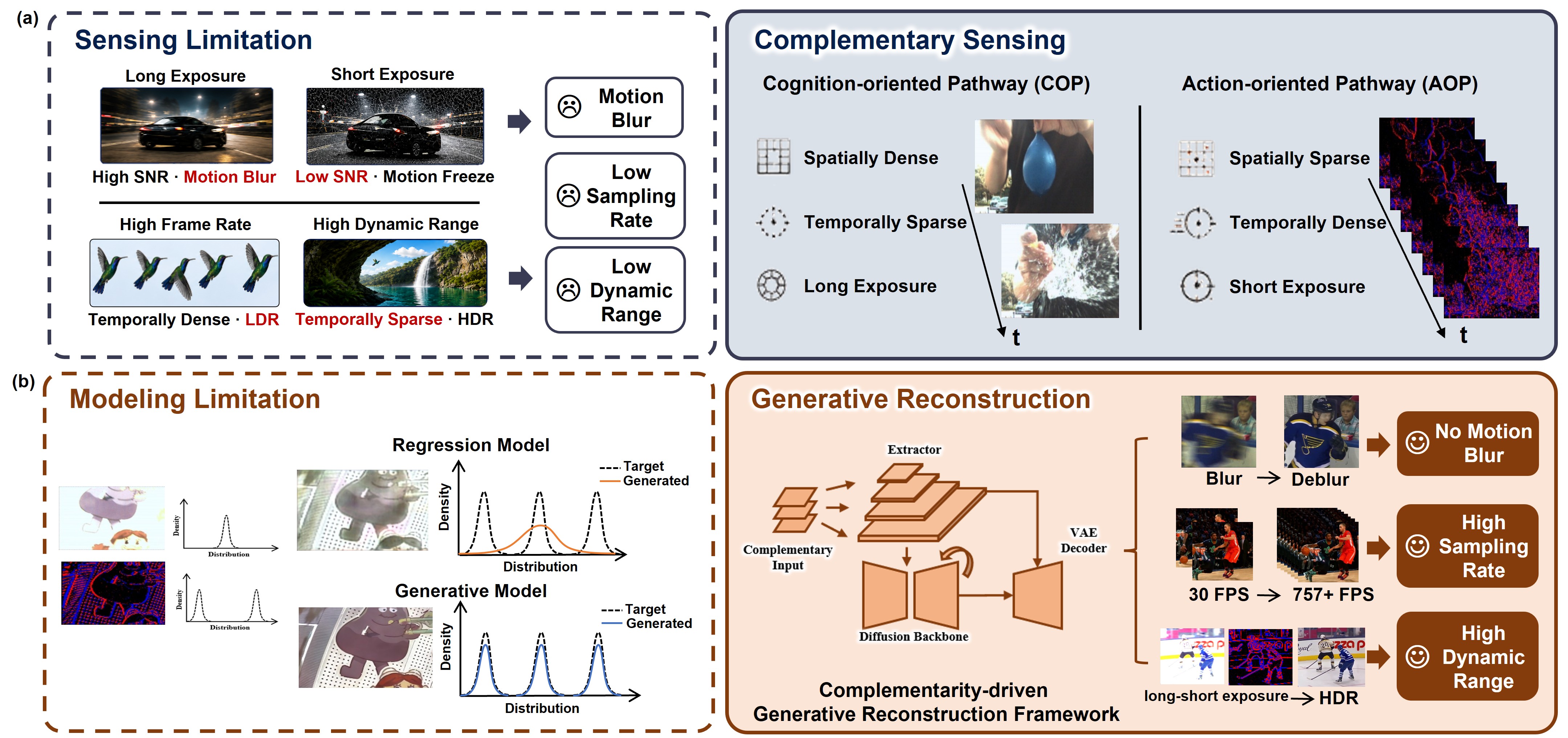}
    \caption{\textbf{Motivation for capturing extreme dynamic scenes with the Complementary Vision Sensor (CVS) and reconstructing them using our complementarity-driven generative reconstruction framework, CoGenRec.}
    (a) Limitations of conventional dense-sampling image sensors and how CVS alleviates them. Under limited bandwidth and power, frame rate and dynamic range are mutually constrained, while signal-to-noise ratio (SNR) and motion blur involve an inherent trade-off. CVS alleviates these constraints through two synergistic visual pathways.
    (b) Regression-based reconstruction versus CoGenRec. In the ill-posed solution space induced by extreme scenes, regression models tend to average plausible solutions, producing over-smoothed results. CoGenRec instead exploits the spatial, temporal, and exposure complementarity of CVS to guide generative sampling and jointly reconstruct blur-free, high-frame-rate HDR-fused video.}
    \label{fig:motivation}
\end{figure*}

\IEEEraisesectionheading{\section{Introduction}\label{sec:introduction}}
\IEEEPARstart{R}{ecording} and reconstructing fast motion under challenging illumination is essential for faithful visual recording and reliable analysis in machine vision, computational photography, and scientific observation~\cite{vidar,wang2021deep,liang2020single}. 
Such scenes demand simultaneous capture of rapid temporal dynamics, fine spatial details, and a wide dynamic range. However, conventional single-stream, densely sampled sensors face inherent acquisition trade-offs, as illustrated in Fig.~\ref{fig:motivation}(a): (i) under practical readout-bandwidth and power constraints~\cite{shu2017real,tabrizchi2022ocelli}, there is a trade-off between frame rate and quantization precision at a fixed spatial resolution: higher frame rates limit the available bit depth per pixel, thereby constraining the representable dynamic range of each frame; and (ii) exposure time governs a trade-off between photon collection and motion fidelity. 
Long exposures improve the signal-to-noise ratio (SNR) in low-light conditions, but introduce motion blur and can cause saturation in bright regions, whereas short exposures reduce motion blur and preserve highlight details at the expense of low-light SNR. 
Consequently, high-contrast fast-moving scenes often contain a combination of temporally undersampled motion, exposure-induced blur, and clipped intensity information. 
Learning-based restoration methods can mitigate these degradations using learned visual priors~\cite{nah2017deep,oh2022demfi,kalantari2017deep,li2022all}; however, recovering fine details, rapid temporal dynamics, and a wide range of intensities simultaneously remains challenging when the captured measurements provide insufficient evidence about the underlying scene. These limitations motivate a different approach to visual acquisition that captures complementary information at the hardware level.


The brain-inspired Complementary Vision Sensor (CVS), known as Tianmouc~\cite{Tianmouc}, introduces such an approach through two synchronized sensing pathways with distinct sampling characteristics. Its cognition-oriented pathway (COP) produces spatially dense RGB frames at 30 frames per second (FPS), while its action-oriented pathway (AOP) captures multi-bit spatial-difference (SD) and temporal-difference (TD) signals at 757–10,000 FPS. 
Rather than relying on a single stream to capture all aspects of a dynamic scene, CVS distributes visual information across complementary measurements. The RGB pathway provides dense spatial detail and color information, while the high-speed differential pathway supplies high-temporal-resolution structural cues and temporal changes between RGB exposures. Their complementary exposure characteristics further provide information that can support reconstruction across a wider range of scene intensities. 
This sensing paradigm creates an opportunity to reconstruct high-quality video by combining spatially rich appearance observations with high-speed motion and structural cues. Realizing this opportunity, however, requires reconstruction methods that effectively exploit the distinct yet interdependent information carried by the two pathways. 

Recent studies have explored CVS reconstruction in several settings, including general RGB synthesis~\cite{Tianmouc,lin2026brain}, HDR video propagation~\cite{lin2026unified}, generative extreme-speed reconstruction~\cite{meng2025diffusion}, and RGB deblurring~\cite{meng2026deblur}. 
These efforts establish an important foundation but leave the full potential of CVS complementarity underexplored. 
Self-supervised methods~\cite{Tianmouc,lin2026brain} reduce dependence on large collections of paired ground-truth videos, but the lack of direct high-quality target supervision can limit reconstruction fidelity. Moreover, most existing methods~\cite{Tianmouc,lin2026brain,lin2026unified,meng2026deblur} also rely on regression-based reconstruction, which tends to produce average predictions when fine textures and appearance details are insufficiently constrained by the measurements. The resulting reconstructions can exhibit over-smoothed structures and muted colors, particularly when motion blur and exposure clipping obscure visual information, as illustrated in Fig.~\ref{fig:motivation}(b). Generative models offer a promising alternative by leveraging learned video priors to recover missing visual details. However, existing approaches have not fully integrated large-scale pretrained video priors with the complementary constraints of CVS. For example, CBRDM~\cite{meng2025diffusion} applied diffusion models to joint deblurring and interpolation, but concatenates differential signals into a scratch-trained network, without exploiting large-scale pretrained video priors or addressing HDR recovery. 
Consequently, a unified framework that exploits CVS complementarity while integrating powerful generative priors remains lacking.
Here, we present \emph{CoGenRec}, a complementarity-driven generative reconstruction framework to recover blur-free, high-frame-rate, and HDR-fused video from CVS complementary measurements. Our central insight is that high-fidelity reconstruction requires combining the sensor-grounded evidence with the learned priors of a video generation model. 
Sensor measurements provide scene constraints that preserve color fidelity, structural clarity, and temporal consistency, while pretrained video priors help recover appearance details that are weakly observed or lost during acquisition.
CoGenRec integrates these complementary sources of information in a unified architecture that balances reconstruction fidelity with perceptual quality~\cite{blau2018perception}. 
Specifically, it exploits three forms of CVS complementarity.
1) \textbf{Spatio-temporal Sampling Complementarity.} We combine spatially dense but temporally sparse RGB frames with temporally dense but spatially sparse differential signals. The latter capture rapid scene dynamics between RGB exposures, enabling high-frame-rate reconstruction while retaining spatial detail. 
2) \textbf{Complementarity between High-SNR Imaging and Fast-Motion Sensing.} The long-exposure RGB pathway provides high-SNR color and appearance information, while the high-speed differential pathway captures intra-exposure motion and structural edges. Their integration helps recover sharp structures obscured by motion blur in RGB frames.
3) \textbf{Exposure complementarity for dynamic-range extension.} The two pathways operate with independent exposure settings, providing complementary observations across intensity regimes. Long-exposure RGB measurements capture shadow details, while short-exposure differential measurements preserve structural cues in bright regions vulnerable to RGB saturation. These measurements provide additional constraints for HDR fusion~\cite{Tianmouc}.

Building on these principles, CoGenRec integrates multiscale sensor guidance, pretrained generative priors, and detail-preserving decoding into a unified reconstruction pipeline. Its architecture comprises three components. 
First, the Multi-Scale Spatio-Temporal Guidance Extractor converts CVS measurements into dense structural and motion guidance. Its Exposure-Aware Endpoint Encoder extends the multibranch design of STGDNet~\cite{meng2026deblur}, fusing long-exposure RGB with intra-exposure $\mathcal{SD}$/$\mathcal{TD}$ cues to restore motion-blurred and saturated regions at RGB exposure boundaries. A Bidirectional Recurrent Temporal Propagator then uses multi-scale convolutional gated recurrent units (ConvGRUs) and inter-exposure differential measurements to propagate guidance to intermediate timestamps. Second, the Complementarity-Guided Video Diffusion Backbone injects sensor-derived guidance into a pretrained video diffusion model through a multi-scale ControlNet~\cite{zhang2023adding}, using video priors to recover missing appearance details.
Its temporal layers capture inter-frame dependencies, promoting temporal coherence while reducing color drift. 
Third, the Detail-Preserving VAE Decoder reduces information loss caused by latent-space compression. It integrates multiscale guidance features into the decoding hierarchy through skip connections to restore high-frequency details~\cite{rombach2022high}.

CoGenRec is trained in three stages: pretraining the guidance extractor, adapting ControlNet and temporal layers with the spatial generative backbone frozen, and optimizing decoder fusion modules for fine-detail recovery. On SportsSloMo-CVS, it improves Peak Signal-to-Noise Ratio (PSNR) over the strongest general-purpose baseline for each task by 4.40, 3.89, and 5.04 dB for deblurring, interpolation, and HDR fusion, respectively. It also demonstrates strong perceptual quality on 129 real-world scenes captured with real CVS hardware.

\textbf{Main Contributions and Extensions over Preliminary Versions.}
Preliminary versions of this work appeared at IEEE/CVF ICCV 2025~\cite{meng2025diffusion} and CVPR 2026~\cite{meng2026deblur}. Building upon these task-specific studies, this work develops a unified reconstruction framework and substantially expands the task formulation, network architecture, training data, and real-world evaluation. Its main contributions are as follows:

\begin{itemize}
    \item \textbf{Systematic formulation of CVS complementarity:} We characterize and formulate CVS acquisition in terms of three complementary properties: complementary spatio-temporal sampling, the decoupling of high-SNR imaging from fast-motion sensing, and exposure complementarity for dynamic-range extension. This formulation provides a common basis for integrating heterogeneous sensor measurements across reconstruction tasks, extending prior work that exploited CVS cues primarily within individual task settings.
    
    \item \textbf{Unified complementarity-driven generative reconstruction framework:} We introduce CoGenRec, which unifies video deblurring, high-temporal-resolution frame interpolation, and HDR fusion within a sensing-to-reconstruction pipeline. By combining multiscale sensor-grounded guidance, video diffusion priors, and detail-preserving decoding, CoGenRec integrates sensor fidelity with generative recovery of missing visual information.
    
    \item \textbf{Expanded training data and comprehensive real-world validation:} We develop a scalable, physically grounded CVS simulation pipeline incorporating  OpenVid-1M videos~\cite{nan2024openvid}, sensor-noise modeling, and exposure augmentation to construct approximately one million training clips. We further evaluate CoGenRec on 129 challenging scenes captured with real CVS hardware, assessing generalization to unseen scenes under fast motion, severe blur, and extreme lighting contrast.
    
\end{itemize}

\section{Related Works}\label{sec:related_works}

\begin{figure*}[!t]
    \centering
    \includegraphics[width=\textwidth]{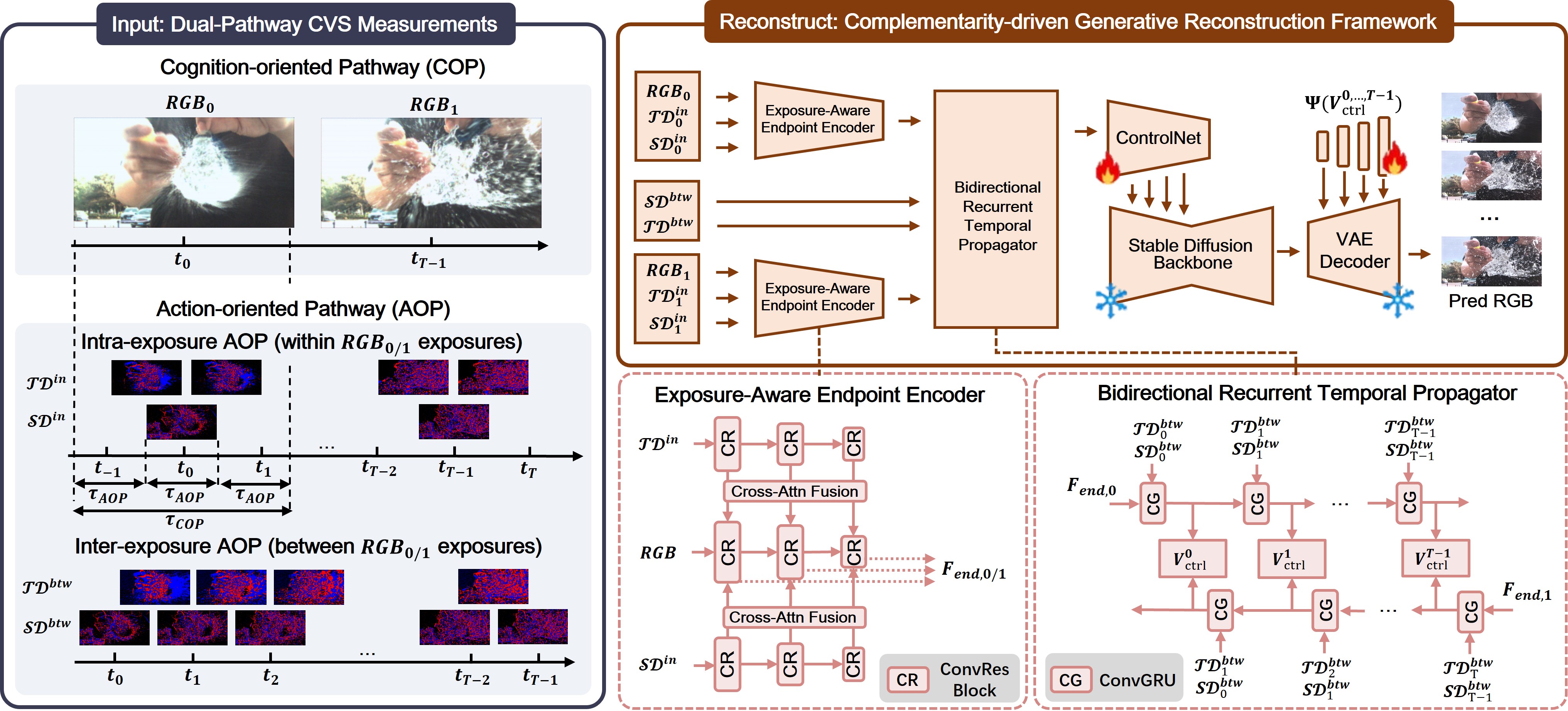}
    \caption{\textbf{Overall architecture of CoGenRec.} The Multi-Scale Spatio-Temporal Guidance Extractor first encodes the dual-pathway CVS measurements into dense multi-scale guidance volumes. These volumes condition the Complementarity-Guided Video Diffusion Backbone through a multi-scale ControlNet.
    Finally, the Detail-Preserving VAE Decoder fuses the uncompressed multi-scale guidance features into the VAE decoding hierarchy and produces sharp, high-frame-rate, and HDR-fused RGB frames.
    }
    \label{fig:overall_framework}
\end{figure*}

\subsection{Brain-Inspired Visual Sensing and CVS Reconstruction}

Conventional single-stream image sensors densely sample images at fixed frame rates, limiting their ability to capture rapid dynamics under limited bandwidth and power budgets. Brain-inspired visual sensors offer alternative sampling paradigms:  event cameras asynchronously record logarithmic intensity changes~\cite{gallego2022event}, whereas spiking cameras encode photon accumulation into dense binary spike streams~\cite{dong2021spiking}. The CVS~\cite{Tianmouc} adopts a distinct approach, synchronously combining dense RGB frames with high-frame-rate, multi-bit SD and TD measurements, providing complementary appearance, structural, and motion cues. 

Existing CVS reconstruction methods exploit different aspects of these measurements. The original Tianmouc study demonstrated general RGB reconstruction by fusing COP and AOP measurements with a U-Net~\cite{Tianmouc}. IGFNet~\cite{lin2026brain} combines  integration, optical-flow, and HDR priors with symmetric self-supervised mutual prediction to handle high-speed motion and HDR scenes, but relies on proxy-based supervision rather than high-quality target videos. PFANet~\cite{lin2026unified} uses COP frames as absolute-radiance anchors and propagates them to high-rate target timestamps using AOP gradients and motion cues for HDR video reconstruction, without jointly restoring blurred RGB endpoints. CBRDM~\cite{meng2025diffusion} applies diffusion models to extreme-speed sRGB video reconstruction but trains its diffusion backbone from scratch, without leveraging large-scale pretrained video priors or addressing HDR fusion.  STGDNet~\cite{meng2026deblur} explicitly exploits intra-exposure difference cues for high-fidelity RGB image deblurring, but focuses on endpoint restoration rather than unified video reconstruction. Despite these advances, existing CVS methods largely address general fusion, HDR imaging, extreme-speed reconstruction, and deblurring separately. A unified generative framework that jointly restores blurred or saturated RGB endpoints, synthesizes intermediate frames, and reconstructs high-frame-rate HDR video while exploiting both CVS measurements and pretrained video priors remains underexplored.

\subsection{Task-Unified Dynamic-Scene Reconstruction}
Dynamic scenes can exhibit three coupled degradations: exposure integration causes motion blur, temporal undersampling omits intermediate frames, and radiometric clipping destroys intensity information. Recovering sharp, high-frame-rate HDR video therefore requires jointly addressing spatial detail, temporal resolution, and dynamic range. 

DeMFI~\cite{oh2022demfi} jointly deblurs low-frame-rate observations and interpolates intermediate frames to produce sharp, high-frame-rate video. Event-guided approaches exploit intra-exposure motion cues to reconstruct sharp high-frame-rate videos from blurry inputs~\cite{pan2019bringing,sun2023eventinterp,kim2025unified}. For spiking-based reconstruction, by contrast, Chang \etal~\cite{chang20231000} combine a spiking camera with an alternating-exposure RGB camera and use spike-derived motion constraints to recover HDR color video at 1000~FPS. 
These methods typically couple deblurring with frame interpolation or high-frame-rate reconstruction with HDR imaging, but do not establish a unified solution for simultaneous deblurring, intermediate-frame synthesis, and HDR recovery as a single reconstruction process.

\subsection{Diffusion Priors for Conditional Video Restoration}

Diffusion models provide powerful generative priors for recovering visual context under incomplete or degraded observations, with applications to image deblurring~\cite{whang2022deblurring}, video frame interpolation~\cite{danier2024ldmvfi}, and event-guided HDR recovery~\cite{yang2025event}. Latent Diffusion Models~\cite{rombach2022high} improve computational efficiency by performing generation in a compressed latent space, while conditional control~\cite{zhang2023adding} and video diffusion architectures~\cite{blattmann2023align} enable pre-trained generative models to incorporate external observations and temporal relationships. 
However, plausible generation alone does not guarantee fidelity to measured scene content, temporal consistency, or preservation of fine details.

CVS provides explicit, complementary physical measurements that can constrain the reconstruction. RGB frames and intra-exposure spatio-temporal difference signals provide observations at RGB exposure boundaries, whereas inter-exposure differential measurements supply additional motion cues for inferring intermediate frames. CoGenRec exploits this structure by conditioning a pretrained video diffusion model on dense multi-scale CVS representations. This integration combines learned appearance priors with sensor-grounded structural and temporal constraints, enabling unified reconstruction of sharp, high-frame-rate video with HDR fusion.

\section{Method}

\subsection{Preliminary: Complementary Vision Sensor}
\label{sec:cvs_model}

The CVS captures scene dynamics through two synergistic vision pathways with complementary sampling and integration mechanisms. The COP records spatially dense RGB frames, while the AOP captures high-frame-rate spatial and temporal difference signals. Their joint measurements provide complementary constraints on appearance, structure, and motion, forming the physical foundation of CoGenRec.

The COP produces spatially dense RGB frames at approximately 30~FPS with 10-bit quantization, yielding $\mathbf{RGB} \in \mathbb{R}^{H \times W \times 3}$.
Its relatively long exposure time $\tau_{cop}$ accumulates sufficient photons to capture high-fidelity color textures. 
Let $\mathbf{I}^{\mathrm{RGB}}_t \in \mathbb{R}^{H \times W \times 3}$ denote the latent instantaneous RGB intensity map at time $t$, and let $t_k$ denote the exposure midpoint of the $k$-th frame. A captured frame $\mathbf{RGB}_k$ is modeled as the temporal integration of this intensity map over the exposure interval, followed by 10-bit quantization denoted by $\sigma_{\mathrm{cop}}(\cdot)$:
\begin{equation}
\mathbf{RGB}_k =
\sigma_{\mathrm{cop}}\!\left(
\int_{t_k-\tau_{\mathrm{cop}}/2}^{t_k+\tau_{\mathrm{cop}}/2}
\mathbf{I}^{\mathrm{RGB}}_t\,dt
\right).
\end{equation}
The long exposure improves photon collection and preserves low-light shadow details, but can introduce motion blur as well as saturated pixels in bright regions.

The AOP captures high-speed spatial and temporal differences at 757--10,000~FPS with configurable quantization precision ranging from $\pm$7-bit to $\pm$1-bit. This pathway senses single-channel intensity without color information. Its outputs comprise spatial differences $\mathcal{SD} \in \mathbb{R}^{H \times W \times 2}$ and temporal differences $\mathcal{TD} \in \mathbb{R}^{H \times W \times 1}$, where the two spatial-difference channels correspond to two spatial directions.
The AOP integrates incident light over its own ultra-short exposure time $\tau_{\mathrm{aop}}$. Let $\{t_n\}$ denote the centers of successive exposures, where $t_n-t_{n-1}=\tau_{\text{diff}}$. Let $\mathbf{I}^{\mathrm{AOP}}_s \in \mathbb{R}^{H \times W}$ denote the latent instantaneous AOP intensity map at time $s$, and let $\mathbf{L}_t$ denote the integrated light intensity over a single exposure centered at $t$:
\begin{equation}
\mathbf{L}_t =
\int_{t-\tau_{\mathrm{aop}}/2}^{t+\tau_{\mathrm{aop}}/2}
\mathbf{I}^{\mathrm{AOP}}_s \, ds.
\end{equation}

At each sampling instant $t=t_n$, spatial differences are formed from the spatial gradients of $\mathbf{L}_t$ along the $\pm45^{\circ}$ directions induced by the cross-pixel architecture of the CVS~\cite{Tianmouc}, while temporal differences are formed by subtracting the integrated intensities of two consecutive exposures separated by $\tau_{\text{diff}}$. Both undergo multi-bit quantization, denoted by $\sigma_{\mathrm{aop}}(\cdot)$, with precision determined by the configured sampling rate. $[\cdot\,;\,\cdot]$ denotes channel-wise concatenation:
\begin{equation}
\begin{split}
\mathcal{SD}_t &= \sigma_{\mathrm{aop}}\!\left(
\left[
\nabla_{+45^{\circ}} \mathbf{L}_t \,;\,
\nabla_{-45^{\circ}} \mathbf{L}_t
\right]\right), \\
\mathcal{TD}_t &= \sigma_{\mathrm{aop}}\!\left(
\mathbf{L}_t - \mathbf{L}_{t-\tau_{\text{diff}}}
\right).
\end{split}
\end{equation}
The AOP’s short exposure limits motion blur and highlight saturation, retaining structural and motion cues in rapidly moving or brightly illuminated regions. 
They complement the COP’s color and appearance information with high-temporal-resolution structural and motion cues. Together, the two pathways provide complementary constraints for reconstructing visual information obscured by motion blur, temporal undersampling and exposure saturation.

\subsection{Problem Formulation}
\label{sec:problem_formulation}

CoGenRec formulates reconstruction of extreme dynamic scenes as a joint reconstruction problem conditioned on dual-pathway CVS measurements, aiming to recover sharp, high-frame-rate, and well-exposed RGB video. The underlying premise is that the two pathways retain complementary scene information under rapid motion and extreme illumination, jointly providing evidence for recovering visual content that neither pathway alone can fully capture. Following the direct RGB-generation strategy of UltraFusion~\cite{chen2025ultrafusion}, CoGenRec generates displayable RGB frames supervised by well-exposed RGB frames from natural-image data, without explicitly reconstructing intermediate linear radiance or applying a separate tone-mapping stage.


\subsubsection{Input Organization}
CoGenRec organizes the dual-pathway measurements into two temporally aligned streams according to their acquisition intervals: an intra-exposure stream for RGB endpoint deblurring and HDR fusion, and an inter-exposure stream for intermediate high-frame-rate frame reconstruction. This separation allows the reconstruction network to exploit the distinct roles of measurements acquired within and between consecutive RGB exposures.

\paragraph{Intra-exposure stream ($\mathcal{X}^{\mathrm{in}}$)}
For each long-exposure frame $\mathbf{RGB}_k$, we collect the $\mathcal{TD}$ measurements sampled throughout its exposure window, together with the $\mathcal{SD}$ measurement sampled at the RGB exposure midpoint:
\begin{equation}
\mathcal{X}_{k}^{\mathrm{in}}
=
\left\{
\left(
\mathcal{TD}_{m}^{\mathrm{in}}
\right)_{t_m \in
[t_k-\tau_{cop}/2,\; t_k+\tau_{cop}/2]},
\;
\mathcal{SD}_{t_k}^{\mathrm{in}}
\right\}.
\end{equation}
Here, $\mathcal{SD}_{t_k}^{\mathrm{in}}$ denotes the spatial-difference measurement sampled at the midpoint $t_k$ of the RGB exposure. The $\mathcal{TD}_{m}^{\mathrm{in}}$ sequence provides high-temporal-resolution motion cues throughout the exposure for restoring potentially blurred RGB frames. 
Meanwhile, the midpoint $\mathcal{SD}_{t_k}^{\mathrm{in}}$ measurement supplies complementary structural information captured with a short exposure, particularly in regions where the long-exposure RGB frame may be saturated. Together, these measurements provide complementary constraints for recovering sharp details and fusing dynamic range information at the RGB endpoint.

\paragraph{Inter-exposure stream ($\mathcal{X}^{\mathrm{btw}}$)}

Given two consecutive RGB observations, $\mathbf{RGB}_0$ and $\mathbf{RGB}_1$, CoGenRec organizes the high-speed differential measurements between their exposure midpoints into an inter-exposure stream. Let $T$ denote the number of SD sampling instants in this stream, indexed by $n=0,\ldots,T-1$. The first and last spatial-difference measurements, $\mathcal{SD}_0^{\mathrm{btw}}$ and $\mathcal{SD}_{T-1}^{\mathrm{btw}}$, are synchronized with the exposure midpoints of $\mathbf{RGB}_0$ and $\mathbf{RGB}_1$, respectively.
The $T-1$ intervals between successive SD sampling instants correspond to $\{\mathcal{TD}_n^{\mathrm{btw}}\}_{n=1}^{T-1}$. To organize the inputs consistently for bidirectional propagation, we retain two additional boundary TD entries, corresponding to the intervals from the first RGB exposure midpoint to the first SD sampling instant and from the last SD sampling instant to the second RGB exposure midpoint. Since the corresponding timestamps coincide at each boundary, both entries are set to zero. The inter-exposure stream therefore comprises a TD sequence of length $T+1$ and $T$ SD measurements:
\begin{equation}
\begin{split}
\mathcal{X}^{\mathrm{btw}} &= \left(
\left\{\mathcal{TD}_n^{\mathrm{btw}}\right\}_{n=0}^{T},\,
\left\{\mathcal{SD}_n^{\mathrm{btw}}\right\}_{n=0}^{T-1}
\right), \\
\mathcal{TD}_0^{\mathrm{btw}} &= \mathcal{TD}_T^{\mathrm{btw}} = \mathbf{0}.
\end{split}
\end{equation}

The inter-exposure TD measurements describe temporal intensity changes, while the SD measurements provide complementary spatial-structure cues. Together, they characterize scene evolution between the two RGB exposure midpoints and provide motion and structure guidance for intermediate-frame reconstruction.

The complete input to CoGenRec is therefore:
\begin{equation}
\mathcal{X} = \left\{ \mathbf{RGB}_0,\, \mathbf{RGB}_1,\, \mathcal{X}_{0}^{\mathrm{in}},\, \mathcal{X}^{\mathrm{btw}},\, \mathcal{X}_{1}^{\mathrm{in}} \right\}.
\end{equation}
This organization preserves the distinction between measurements that constrain the restoration of exposure-boundary frames and those that guide temporal reconstruction between them, while allowing both streams to contribute to a unified output.

\subsubsection{Unified Reconstruction Output}

Given the multi-modal input $\mathcal{X}$, CoGenRec aims to reconstruct extreme dynamic scenes as a sharp, high-frame-rate, and well-exposed RGB sequence:
\begin{equation}
\hat{\mathcal{S}}
=
\left\{
\hat{\mathbf{S}}_t
\right\}_{t=0}^{T-1},
\qquad
\hat{\mathbf{S}}_t \in \mathbb{R}^{H \times W \times 3},
\end{equation}
where $\hat{\mathbf{S}}_t$ denotes a sharp, well-exposed RGB frame corresponding to the sampling instant of $\mathcal{SD}_t^{\mathrm{btw}}$, and $t=0,\ldots,T-1$ is the discrete frame index. The reconstruction is modeled as:
\begin{equation}
\hat{\mathcal{S}}
=
\mathcal{G}_{\theta}
\left(
\mathbf{z}
\mid
\mathcal{X},
\mathbf{c}_{\emptyset}
\right),
\end{equation}
where $\mathcal{G}_{\theta}$ denotes the conditional generative reconstruction model parameterized by $\theta$, $\mathbf{z}$ is the initial latent noise, and $\mathbf{c}_{\emptyset}$ is the fixed empty-text embedding used by the pretrained diffusion model.

The output sequence includes reconstructed frames at both RGB exposure midpoints and the intermediate timestamps, jointly recovering spatial detail, temporal variation, and appearance in HDR regions through a unified conditional generation process.


\subsection{Network Architecture}
\label{sec:network_architecture}
To realize the unified reconstruction objective, CoGenRec couples multi-scale constraints extracted from CVS measurements with a pre-trained video generative prior. As illustrated in Fig.~\ref{fig:overall_framework}, the architecture comprises three main components: a Multi-Scale Spatio-Temporal Guidance Extractor, a Complementarity-Guided Video Diffusion Backbone, and a Detail-Preserving VAE Decoder.


\subsubsection{Multi-Scale Spatio-Temporal Guidance Extractor}
The temporally aligned COP and AOP measurements provide information about endpoint appearance, intra-exposure degradation, and inter-exposure dynamics. The Multi-Scale Spatio-Temporal Guidance Extractor encodes these signals into a unified set of multiscale features that constrain reconstruction at both observed exposure boundaries and intermediate timestamps. It consists of an Exposure-Aware Endpoint Encoder and a Bidirectional Recurrent Temporal Propagator.

\paragraph{Exposure-Aware Endpoint Encoder}
The endpoint encoder restores feature representations at the two RGB exposure boundaries, where long-exposure observations may contain motion blur and saturated regions. For the left and right endpoints $t=0$ and $t=T-1$, it fuses the corresponding RGB frames, $\mathbf{RGB}_0$ and $\mathbf{RGB}_1$, with their synchronized intra-exposure difference streams, $\mathcal{X}_{0}^{\mathrm{in}}$ and $\mathcal{X}_{1}^{\mathrm{in}}$, respectively.


Denoted by $\mathcal{E}_{end}$, the encoder builds upon the multi-branch design of STGDNet~\cite{meng2026deblur}, while omitting its recurrent modules. 
It comprises three hierarchical encoding blocks, each containing two $3\times3$ residual convolutions and stride-2 downsampling, followed by two successive channel-attention modules that fuse intra-exposure $\mathcal{TD}$ and $\mathcal{SD}$ features into the RGB branch. 
Together with the full-resolution features, the encoder produces four feature scales at progressively reduced spatial resolutions.

For each scale $l\in\{1,\ldots,4\}$, the two endpoint features are formulated as
\begin{equation}
\begin{aligned}
\mathbf{F}_{end,0}^{l}
&=
\mathcal{E}_{end}^{l}
\left(
\mathbf{RGB}_0,\,
\mathcal{X}_{0}^{\mathrm{in}}
\right), \\
\mathbf{F}_{end,1}^{l}
&=
\mathcal{E}_{end}^{l}
\left(
\mathbf{RGB}_1,\,
\mathcal{X}_{1}^{\mathrm{in}}
\right).
\end{aligned}
\end{equation}
Here, $\mathcal{E}_{end}^{l}(\cdot)$ denotes the mapping from the multimodal endpoint inputs to the fused feature at the $l$-th scale. The endpoint feature sets $\{\mathbf{F}_{end,0}^{l}\}_{l=1}^{4}$ and $\{\mathbf{F}_{end,1}^{l}\}_{l=1}^{4}$ constitute multi-scale representations of the two endpoints, integrating boundary RGB appearance with complementary intra-exposure AOP cues and providing reliable anchors for subsequent bidirectional temporal propagation.

\paragraph{Bidirectional Recurrent Temporal Propagator}
Endpoint features constrain the observed exposure boundaries but do not, by themselves, describe the scene's evolution at intermediate timestamps. The Bidirectional Recurrent Temporal Propagator extends these features throughout the output sequence using inter-exposure AOP measurements, providing temporally aligned guidance for high-frame-rate reconstruction.
The inter-exposure AOP encoder, $\mathcal{E}_{aop}$, processes the predefined inter-exposure stream $\mathcal{X}^{\mathrm{btw}}$ through separate $\mathcal{TD}$ and $\mathcal{SD}$ branches. Each branch comprises three hierarchical residual encoding blocks with two $3\times3$ convolutions and stride-2 downsampling per block. Including the full-resolution features, each branch outputs features at four scales whose spatial resolutions match those of the corresponding endpoint features.

Let $\mathcal{E}_{\mathrm{TD}}^l$ and $\mathcal{E}_{\mathrm{SD}}^l$ denote the $\mathcal{TD}$ and $\mathcal{SD}$ branches of the inter-exposure AOP encoder $\mathcal{E}_{aop}$ at scale $l$, respectively. 
For $l\in\{1,\ldots,4\}$ and $t=0,\ldots,T-1$, let $\mathbf{A}_{t,\mathrm{fwd}}^l$ and $\mathbf{A}_{t,\mathrm{bwd}}^l$ denote the AOP input features for forward and backward ConvGRU propagation, respectively, at scale $l$ and time $t$. They are constructed by concatenating the corresponding encoded $\mathcal{TD}$ and $\mathcal{SD}$ features along the channel dimension:
\begin{equation}
\begin{aligned}
\mathbf{A}_{t,\mathrm{fwd}}^l
&=\left[
\mathcal{E}_{\mathrm{TD}}^l\!\left(\mathcal{TD}_t^{\mathrm{btw}}\right)\,;\,
\mathcal{E}_{\mathrm{SD}}^l\!\left(\mathcal{SD}_t^{\mathrm{btw}}\right)
\right],\\
\mathbf{A}_{t,\mathrm{bwd}}^l
&=\left[
\mathcal{E}_{\mathrm{TD}}^l\!\left(\mathcal{TD}_{t+1}^{\mathrm{btw}}\right)\,;\,
\mathcal{E}_{\mathrm{SD}}^l\!\left(\mathcal{SD}_t^{\mathrm{btw}}\right)
\right].
\end{aligned}
\end{equation}
Thus, at the same output timestamp $t$, forward and backward propagation use $\mathcal{TD}_t^{\mathrm{btw}}$ and $\mathcal{TD}_{t+1}^{\mathrm{btw}}$, respectively, while sharing $\mathcal{SD}_t^{\mathrm{btw}}$.
Let $\vec{\mathbf{h}}_t^l,\overleftarrow{\mathbf{h}}_t^l \in \mathbb{R}^{C_l \times H_l \times W_l}$ denote the forward and backward hidden states at scale $l$ and time $t$, respectively, where $C_l$ is the hidden-state channel count and $H_l$ and $W_l$ are the spatial dimensions. The hidden states are initialized from the corresponding endpoint features:
\begin{equation}
\vec{\mathbf{h}}_{-1}^l=\mathbf{F}_{end,0}^l,
\qquad
\overleftarrow{\mathbf{h}}_T^l=\mathbf{F}_{end,1}^l.
\end{equation}
The indices $-1$ and $T$ label only the recurrence boundary states rather than output frames. The forward branch then updates in the order $t=0,\ldots,T-1$, and the backward branch in the order $t=T-1,\ldots,0$:
\begin{equation}
\begin{aligned}
\vec{\mathbf{h}}_t^l
&=\operatorname{ConvGRU}_{\mathrm{fwd}}^l
\left(\mathbf{A}_{t,\mathrm{fwd}}^l,\,\vec{\mathbf{h}}_{t-1}^l\right),\\
\overleftarrow{\mathbf{h}}_t^l
&=\operatorname{ConvGRU}_{\mathrm{bwd}}^l
\left(\mathbf{A}_{t,\mathrm{bwd}}^l,\,\overleftarrow{\mathbf{h}}_{t+1}^l\right),
\end{aligned}
\end{equation}
here, $\operatorname{ConvGRU}_{\mathrm{fwd}}^l$ and $\operatorname{ConvGRU}_{\mathrm{bwd}}^l$ denote the forward and backward ConvGRU update operators at scale $l$, respectively.

At each scale $l$, we concatenate the forward and backward hidden states channel-wise at each timestamp and arrange the resulting features along the temporal dimension to form a spatio-temporal control volume:
\begin{equation}
\mathbf{V}_{ctrl}^l
=
\left(
\left[\vec{\mathbf{h}}_t^l\,;\,
\overleftarrow{\mathbf{h}}_t^l\right]
\right)_{t=0}^{T-1}
\in \mathbb{R}^{2C_l \times T \times H_l \times W_l}.
\end{equation}

The four control volumes, $\mathbf{V}_{ctrl}=\{\mathbf{V}_{ctrl}^l\}_{l=1}^{4}$, together form a unified multi-scale guidance representation with high temporal resolution that encodes scene structure and its temporal evolution throughout the output sequence. They are shared by the Complementarity-Guided Video Diffusion Backbone and the Detail-Preserving VAE Decoder, providing measurement-grounded guidance for both generative reconstruction and fine-detail recovery.

\subsubsection{Complementarity-Guided Video Diffusion Backbone}
The guidance extractor provides observation-grounded structural and motion constraints, but some appearance information remains weakly observed or unavailable in the sensor measurements. CoGenRec uses pretrained diffusion priors to recover plausible visual details, while CVS-derived guidance maintains fidelity to the observed scene.

We construct the video UNet $\mathcal{U}_{vid}$ by incorporating AnimateDiff motion modules~\cite{guo2024animatediff} into the image UNet of a pretrained LDM~\cite{rombach2022high}, and inject CVS guidance through a ControlNet-based conditioning network $\mathcal{C}_{net}$~\cite{zhang2023adding}.
$\mathcal{C}_{net}$ use a multiscale embedder $\mathcal{E}_{ms}$, which is adapted from the pyramid fusion architecture of Yang et al.~\cite{yang2025event}, aggregates the four-scale guidance collection $\mathbf{V}_{ctrl}$ into a conditioning feature $\widetilde{\mathbf{V}}_{ctrl}$. At diffusion timestep $k$, let $\mathbf{z}_k$ denote the current noisy video latent. Conditioned on the aggregated feature, $\mathcal{C}_{net}$ produces a collection of control residuals $\mathbf{R}_{ctrl}$ for injection into the downsampling and middle layers of the video UNet. The video UNet uses these residuals to predict the noise $\widehat{\boldsymbol{\epsilon}}_{\theta}$:
\begin{equation}
\begin{aligned}
\widetilde{\mathbf{V}}_{ctrl}
&= \mathcal{E}_{ms}\!\left(\mathbf{V}_{ctrl}\right), \\
\mathbf{R}_{ctrl}
&= \mathcal{C}_{net}\!\left(
\mathbf{z}_k,\, k,\, \mathbf{c}_{\emptyset},\,
\widetilde{\mathbf{V}}_{ctrl}
\right), \\
\widehat{\boldsymbol{\epsilon}}_{\theta}
&= \mathcal{U}_{vid}\!\left(
\mathbf{z}_k,\, k,\, \mathbf{c}_{\emptyset}
\,;\, \mathbf{R}_{ctrl}
\right).
\end{aligned}
\end{equation}

Each control residual is generated through a zero-initialized output convolution and added to the corresponding video UNet feature. The predicted noise is used by the denoising sampler to update the video latent. During training, the pretrained spatial UNet remains frozen, while the ControlNet and video motion modules are updated.

This design distinguishes generative completion from unconstrained synthesis: the pretrained model supplies learned appearance knowledge, whereas the complementary sensor measurements provide scene-specific evidence. The designed architecture is intended to reduce structural drift and temporal inconsistency while improving the recovery of textures and colors that cannot be reliably inferred from degraded RGB frames alone.

\subsubsection{Detail-Preserving VAE Decoder}

Latent diffusion models perform generation in a compressed representation, improving computational efficiency but potentially attenuating fine spatial structures. For reconstruction tasks that require accurate textures and edges, relying exclusively on the decoded latent representation may therefore limit detail fidelity.

CoGenRec introduces a Detail-Preserving VAE Decoder that integrates the four-scale guidance volumes $\mathbf{V}_{ctrl}$ derived from CVS observations into the VAE decoding hierarchy through skip connections. These guidance features bypass VAE latent-space compression and provide additional high-resolution structural information during image reconstruction, supporting the recovery of fine spatial details.

Specifically, we introduce a residual spatial fusion module before each of the four VAE decoder blocks. Decoding proceeds from coarse to fine, corresponding to guidance scales $l=4,3,2,1$, with each fusion module receiving guidance features at the same spatial resolution as the current decoder feature.

Fusion is performed frame by frame. Let $\mathbf{V}_{ctrl,t}^{l}$ denote the temporal slice of $\mathbf{V}_{ctrl}^{l}$ at timestamp $t$, and let $\widetilde{\mathbf{H}}_{dec,t}^{l}$ denote the VAE decoder feature before fusion at the corresponding spatial scale. For $l\in\{1,\ldots,4\}$ and $t=0,\ldots,T-1$, the fused feature $\overline{\mathbf{H}}_{dec,t}^{l}$ is given by
\begin{equation}
\overline{\mathbf{H}}_{dec,t}^{l}
=
\widetilde{\mathbf{H}}_{dec,t}^{l}
+
\Phi_l\!\left(
\left[
\Psi_l\!\left(\mathbf{V}_{ctrl,t}^{l}\right)\,;\,
\widetilde{\mathbf{H}}_{dec,t}^{l}
\right]
\right).
\end{equation}

Here, $\Psi_l$ is a convolutional adaptation mapping that matches the guidance channels to those of the current decoder feature, and $\Phi_l$ comprises two residual blocks that produce a detail-compensation feature of the same shape as the decoder feature. The fused feature $\overline{\mathbf{H}}_{dec,t}^{l}$ is subsequently processed by the corresponding original VAE decoder block. During training, the original VAE remains frozen, and only the newly introduced fusion modules are updated.

\subsection{Three-Stage Training Strategy}
\label{sec:training_strategy}

CoGenRec adopts a three-stage training strategy to progressively learn observation-grounded guidance, conditional generation, and detail recovery while retaining the pretrained appearance prior.

\subsubsection{Stage 1: Training the Guidance Extractor}
The Multi-Scale Spatio-Temporal Guidance Extractor is trained from scratch to transform complementary CVS measurements into structural and motion guidance, establishing an observation-grounded guidance representation for the subsequent stages.

During training, a lightweight auxiliary decoder is attached to $\mathbf{V}_{ctrl}$ and trained jointly with the guidance extractor to produce an RGB sequence. The predicted sequence is supervised by paired sharp, well-exposed RGB frames using a Charbonnier loss~\cite{charbonnier1997deterministic} and an LPIPS perceptual loss~\cite{zhang2018unreasonable}:
\begin{equation}
\mathcal{L}_{ext}=\mathcal{L}_{char}+0.1\mathcal{L}_{LPIPS}.
\end{equation}

After Stage 1 converges, the auxiliary decoder is removed and the guidance extractor is frozen, retaining only its multi-scale spatio-temporal features for the subsequent stages.

\subsubsection{Stage 2: Adapting the Video Diffusion Backbone}
The second stage adapts the video diffusion backbone to condition generation on the multiscale guidance $\mathbf{V}_{ctrl}$ extracted in Stage-1. We freeze the Stage-1 guidance extractor, VAE, and pre-trained spatial UNet, while updating the multi-scale ControlNet and video motion layers.

We encode the paired sharp, well-exposed ground-truth RGB sequence into a clean latent $\mathbf{z}_0$ using the frozen VAE encoder. At a randomly sampled diffusion timestep $k$, sampled noise $\boldsymbol{\epsilon}$ is added to $\mathbf{z}_0$ according to the forward diffusion process, yielding the noisy latent $\mathbf{z}_k$. The video diffusion backbone receives $\mathbf{z}_k$, $k$, and the fixed empty-text embedding $\mathbf{c}_{\emptyset}$, together with guidance $\mathbf{V}_{ctrl}$ extracted from the corresponding CVS measurements through the ControlNet, and produces the noise prediction $\widehat{\boldsymbol{\epsilon}}_{\theta}$ defined above. Training uses the standard diffusion noise-prediction objective:
\begin{equation}
\mathcal{L}_{LDM}
=
\left\|
\boldsymbol{\epsilon}
-
\widehat{\boldsymbol{\epsilon}}_{\theta}
\right\|_2^2.
\end{equation}

Accordingly, the ControlNet learns to convey CVS structural and motion constraints to the generative process, while the video motion layers model cross-frame consistency.

\subsubsection{Stage 3: Fine-tuning the Detail-Preserving VAE Decoder}
The third stage addresses high-frequency detail loss caused by the VAE latent bottleneck. 
In Stage 3, the frozen Stage-2 model generates a denoised latent $\hat{\mathbf{z}}_0$ through iterative denoising conditioned on the CVS guidance $\mathbf{V}_{ctrl}$. The Detail-Preserving VAE Decoder combines this latent with the same multiscale guidance $\mathbf{V}_{ctrl}$ for RGB reconstruction.

We freeze the guidance extractor, ControlNet, complete video UNet, VAE encoder, and original VAE decoder blocks, and train only the newly introduced spatial fusion modules. The final RGB predictions are supervised by paired sharp, well-exposed RGB frames using the same loss combination as Stage 1:
\begin{equation}
\mathcal{L}_{dec}=\mathcal{L}_{char}+0.1\mathcal{L}_{LPIPS}.
\end{equation}

This fine-tuning allows the decoder to compensate for information lost in the latent space using uncompressed CVS features while preserving the learned generative and temporal priors.



\section{Experiments}

\subsection{Datasets}
\label{sec:datasets}


Training a unified CVS reconstruction model requires large-scale data that capture both the complementary measurement characteristics of the sensor and the diversity of real-world scenes.  
The key challenge lies in converting conventional RGB videos into corresponding CVS measurements while accounting for sensor-specific responses, non-idealities, and exposure settings. Existing conversion approaches fall into two categories: (i) Digital Light Processing (DLP)-based hardware conversion, and (ii) software simulation. Hardware conversion captures the physical sensor’s photoelectronic response and non-idealities but incurs substantially higher conversion cost and time. Software simulation enables efficient, large-scale data generation but may not fully reproduce the behavior of physical sensor response. 
We therefore adopt a hybrid data construction strategy that combines hardware-derived measurements with large-scale software simulation, as detailed below.


\textbf{DLP-based Hardware Simulation.}
Following the DLP-based vision-chip characterization approach~\cite{meng2025technicalreportdmdbasedcharacterization} and the acquisition protocol adopted in~\cite{meng2025diffusion,meng2026deblur}, we employ a Digital Micro-Mirror Device (DMD) and its associated optical system to sequentially project source-video RGB frames onto the CVS pixel plane. Hardware triggering and a fixed optical path ensure temporal synchronization between the projected light and the CVS chip, as well as pixel-level spatial alignment. This process captures authentic sensor responses, including noise, nonlinearities, and other hardware-dependent non-idealities.

Consecutive RGB frames are projected at $\tau_{\text{diff}}$, with each frame transition  synchronized to the AOP exposure signal of the CVS. The CVS records $\mathcal{SD}$ measurements at projected instants and $\mathcal{TD}$ measurements between consecutive instants. Meanwhile, the RGB pathway randomly integrates $N$ consecutive projected frames, yielding authentic RGB measurements with varying degrees of motion blur. This procedure yields paired RGB and differential measurements acquired directly from physical CVS hardware.


\textbf{Software Simulation.} Since DLP-based conversion remains constrained by hardware throughput and limited flexibility in data scaling, we additionally employ a software simulation framework based on prior CVS studies~\cite{Tianmouc,lin2026brain,lin2025csvo}. The simulator reproduces the CVS pixel-sampling geometry and differential measurement mechanism, simulating $\mathcal{SD}$ and $\mathcal{TD}$ signals from RGB frames through spatial and temporal differencing followed by multi-bit threshold quantization. To better approximate physical sensor behavior, we incorporate temporally persistent fixed-pattern noise and time-varying stochastic noise according to measurements and statistical analyses of the CVS differential pathway~\cite{wang2026adaptive}. This enables large-scale conversion while preserving the structural characteristics of CVS measurements and approximating the statistical properties of real sensor readouts.

\textbf{Dataset Composition and Sampling.}
The resulting dataset contains approximately one million clips, including about 380K DLP-converted samples from SportsSloMo~\cite{SportsSlomo}, Ego4D~\cite{grauman2022ego4d}, and OpenVid-1M~\cite{nan2024openvid}, together with 630K software-simulated samples from OpenVid-1M. 2.3K and 9.1K paired samples are reserved for validation and testing, respectively. During training, samples are drawn from four data sources with probabilities of 0.20, 0.25, 0.25, and 0.30, for DLP-converted SportsSloMo-CVS, Ego4D-CVS, OpenVid-1M-CVS, and software-simulated OpenVid-1M-CVS, respectively. Thus, 70\% of the sampling probability is assigned to hardware-converted data, while software simulation increases scene diversity and training-scale coverage. In addition, we collect approximately 500 unlabeled CVS samples from more than 120 real-world sequences to evaluate generalization under challenging conditions, including high dynamic range, high-speed motion, and severe motion blur.

\textbf{Exposure Augmentation.} To enhance HDR fusion capability under various exposure conditions, we introduce an exposure augmentation scheme that simulates highlight saturation in the RGB pathway and weak differential pathway responses in low-light regions. A shared random exposure gain is applied to both RGB endpoints, followed by intensity clipping to simulate highlight saturation. For the differential signals, low-amplitude responses are randomly suppressed using stochastic thresholds, with additional noise injected to mimic under-sampling in dark regions. The sharp RGB ground truth remains unchanged, encouraging the model to recover well-exposed targets by exploiting the complementary exposure characteristics of the two pathways. We refer to the settings with and without this augmentation as \emph{with exposure augmentation} and \emph{without exposure augmentation}, respectively.



\begin{figure*}[!t]
    \centering
    \includegraphics[width=\textwidth]{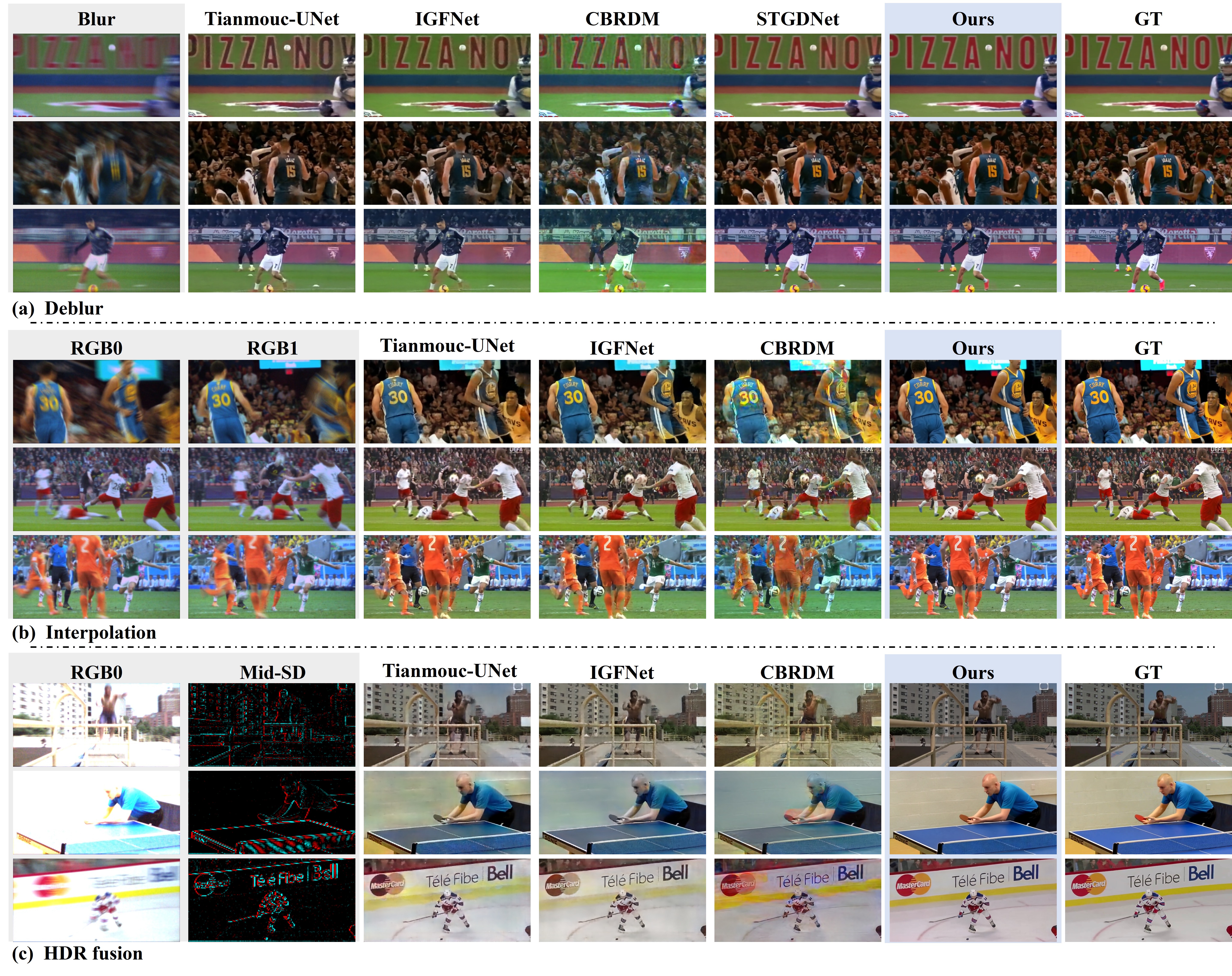}
    \caption{\textbf{Qualitative comparisons on SportsSloMo-CVS under the unified training protocol.} All baselines are retrained on the same large-scale mixed training dataset as our method. Each row shows one test scene. (a) Deblurring results for a blurry RGB endpoint frame. (b) Interpolation results at the temporal midpoint between two adjacent blurry RGB endpoint frames, RGB0 and RGB1. (c) HDR synthesis results at the temporal position corresponding to Mid-SD between two adjacent RGB frames; the highlights in RGB0 are saturated, whereas Mid-SD visualizes the complementary structural cues retained by the short-exposure differential pathway at that time. STGDNet is specific to motion deblurring and is therefore included only in (a).}
    \label{fig:qualitative_comparison}
\end{figure*}

\begin{table*}[!t]
\centering
\tablebodyfont
\setlength{\tabcolsep}{2pt}
\caption{Quantitative motion-deblurring comparison on the SportsSloMo-CVS test set. The blurry RGB inputs integrate 5, 7, 9, and 11 consecutive sharp frames, respectively; Average denotes the arithmetic mean over the four blur levels. The best and second-best results in each column are highlighted in bold and underlined, respectively.}
\begin{tabular}{c|c c c|c c c|c c c|c c c|c c c}
\toprule
\multirow{2}{*}{\rule{0pt}{2.6ex}Method} & \multicolumn{3}{c|}{5} & \multicolumn{3}{c|}{7} & \multicolumn{3}{c|}{9} & \multicolumn{3}{c|}{11} & \multicolumn{3}{c}{Average} \\
\cline{2-16}
& PSNR$\uparrow$ & SSIM$\uparrow$ & LPIPS$\downarrow$ & PSNR$\uparrow$ & SSIM$\uparrow$ & LPIPS$\downarrow$ & PSNR$\uparrow$ & SSIM$\uparrow$ & LPIPS$\downarrow$ & PSNR$\uparrow$ & SSIM$\uparrow$ & LPIPS$\downarrow$ & PSNR$\uparrow$ & SSIM$\uparrow$ & LPIPS$\downarrow$ \\
\midrule
Tianmouc-UNet~\cite{Tianmouc} & 27.50 & 0.8848 & 0.1648 & 27.70 & 0.8906 & 0.1652 & 27.46 & 0.8780 & 0.1844 & 27.11 & 0.8692 & 0.1952 & 27.44 & 0.8807 & 0.1774 \\
IGFNet~\cite{lin2026brain} & 28.09 & 0.9008 & 0.1398 & 28.16 & 0.9038 & 0.1415 & 27.89 & 0.8910 & 0.1583 & 27.43 & 0.8839 & 0.1686 & 27.89 & 0.8949 & 0.1521 \\
CBRDM~\cite{meng2025diffusion} & 22.57 & 0.7730 & 0.1487 & 22.93 & 0.8043 & 0.1503 & 22.43 & 0.7918 & 0.1708 & 22.13 & 0.7711 & 0.1836 & 22.52 & 0.7851 & 0.1633 \\
STGDNet~\cite{meng2026deblur} & \textbf{35.10} & \textbf{0.9598} & \underline{0.0387} & \textbf{34.68} & \textbf{0.9569} & \underline{0.0446} & \textbf{34.12} & \textbf{0.9517} & \underline{0.0523} & \textbf{33.39} & \textbf{0.9463} & \underline{0.0602} & \textbf{34.33} & \textbf{0.9537} & \underline{0.0490} \\
CoGenRec (Ours) & \underline{32.67} & \underline{0.9452} & \textbf{0.0383} & \underline{32.55} & \underline{0.9441} & \textbf{0.0403} & \underline{32.14} & \underline{0.9380} & \textbf{0.0454} & \underline{31.82} & \underline{0.9347} & \textbf{0.0482} & \underline{32.30} & \underline{0.9405} & \textbf{0.0431} \\
\bottomrule
\end{tabular}
\label{tab:comparison_deblur}
\end{table*}

\subsection{Training Details}
\label{sec:training_details}

For the benchmark comparisons in Sec.~\ref{sec:benchmark_comparisons}, CoGenRec reconstructs seven uniformly spaced target frames at a resolution of $320\times640$. Training proceeds in three stages without exposure augmentation: the guidance extractor, video diffusion backbone, and decoder fusion modules are sequentially trained for 100K, 100K, and 30K iterations, respectively, with an initial learning rate of $10^{-4}$ and effective batch sizes of 24, 24, and 8.
We then enable exposure augmentation and initialize the guidance extractor and video diffusion backbone from their counterparts trained without exposure augmentation, fine-tuning each for 100K iterations at an initial learning rate of $10^{-5}$. These modules are subsequently frozen while the decoder fusion modules are trained for 30K iterations at an initial learning rate of $10^{-4}$. This procedure adapts the model to more challenging exposure conditions.

For the higher-frame-rate experiments in Sec.~\ref{sec:higher_frame_rate_scaling}, we further adapt the exposure-augmented seven-frame versions to CVS input modes at 757, 1515, and 3030~FPS through variable-length fine-tuning with mixed sequence lengths of up to 101 frames. The three stages are fine-tuned for 30K, 30K, and 10K iterations at initial learning rates of $10^{-5}$, $10^{-5}$, and $10^{-4}$, respectively, retaining the parameter-freezing strategy of each stage.

All training stages use Adam~\cite{kingma2015adam} with a cosine annealing schedule and 100 warm-up iterations. At inference, the diffusion process uses 50-step DDIM~\cite{song2021denoising} sampling.

\begin{table*}[!t]
\centering
\tablebodyfont
\setlength{\tabcolsep}{3.5pt}
\renewcommand{\arraystretch}{1.10}
\caption{Quantitative frame-interpolation comparison on the SportsSloMo-CVS test set. ``Middle Frame'' denotes the temporal midpoint of the reconstructed sequence, whereas ``All Frames'' denotes all seven reconstructed frames in each sample. PSNR, SSIM, LPIPS, and NIQE are averaged over the corresponding frames and samples, while FID~\cite{heusel2017gans} is computed over the corresponding pooled image set. TR is the motion-compensated temporal residual between adjacent predictions. The best and second-best results in each column are highlighted in bold and underlined, respectively.}
\label{tab:comparison_interpolation}
\begin{tabular}{c|c c c c c|c c c c c|c}
\toprule
\multirow{2}{*}{Method} & \multicolumn{5}{c|}{Middle Frame} & \multicolumn{5}{c|}{All Frames} & \multirow{2}{*}{TR$\downarrow$} \\
\cline{2-11}
& PSNR$\uparrow$ & SSIM$\uparrow$ & LPIPS$\downarrow$ & NIQE$\downarrow$ & FID$\downarrow$
& PSNR$\uparrow$ & SSIM$\uparrow$ & LPIPS$\downarrow$ & NIQE$\downarrow$ & FID$\downarrow$ & \\
\midrule
Tianmouc-UNet~\cite{Tianmouc} & 27.17 & 0.8722 & 0.1835 & 6.56 & 47.86 & 27.32 & 0.8763 & 0.1793 & 6.52 & 33.51 & 0.019735 \\
IGFNet~\cite{lin2026brain} & \underline{27.67} & \underline{0.8888} & \underline{0.1557} & 6.53 & \underline{38.72} & \underline{27.77} & \underline{0.8912} & \underline{0.1541} & 6.53 & \underline{27.05} & \underline{0.018129} \\
CBRDM~\cite{meng2025diffusion} & 21.21 & 0.7429 & 0.2082 & \textbf{5.15} & 62.02 & 21.77 & 0.7624 & 0.1851 & \textbf{5.16} & 35.43 & 0.029459 \\
CoGenRec (Ours) & \textbf{31.18} & \textbf{0.9309} & \textbf{0.0520} & \underline{5.77} & \textbf{13.59} & \textbf{31.66} & \textbf{0.9355} & \textbf{0.0475} & \underline{5.75} & \textbf{6.50} & \textbf{0.016047} \\
\bottomrule
\end{tabular}
\end{table*}

\subsection{Task-specific Benchmark Comparisons}
\label{sec:benchmark_comparisons}
CoGenRec jointly performs video deblurring, frame interpolation, and HDR fusion. We evaluate these three tasks separately to enable fair comparisons with task-specific baselines.

\textbf{Baseline Selection.}
To the best of our knowledge, the selected baselines cover all existing CVS-based reconstruction, frame interpolation, and motion deblurring methods. Tianmouc-UNet, the original U-Net-based reconstruction method introduced with the CVS (Tianmouc) sensor~\cite{Tianmouc}, IGFNet~\cite{lin2026brain}, which fuses complementary intermediate representations through symmetric self-supervised learning, and the diffusion-based CBRDM~\cite{meng2025diffusion} are general CVS reconstruction methods capable of both recovering sharp RGB frames and synthesizing intermediate frames between adjacent RGB exposures. In principle, these methods can support HDR fusion across the two sensing pathways.
STGDNet~\cite{meng2026deblur} is specifically designed for CVS-guided motion deblurring and does not support intermediate-frame reconstruction. Accordingly, we include all four methods above in the deblurring comparison, while using the three general reconstruction methods for the frame interpolation and HDR comparisons. 
PFANet~\cite{lin2026unified} uses linear HDR radiance supervision, which is unavailable in our synthetic training data. We therefore include it only in qualitative comparisons on real HDR scenes.

\textbf{Training Protocol.}
Since the compared methods were originally trained on different data sources, we retrain all baselines in this subsection from their official implementations using the same large-scale mixed training datasets described in Sec.~\ref{sec:datasets} to minimize the influence of training data.

Conventional motion deblurring and frame interpolation generally assume brightness consistency, i.e., the reconstructed frames should preserve the global brightness distribution of the input RGB frames. In contrast, HDR fusion aims to recover well-exposed outputs from measurements with different exposure characteristics; consequently, the global brightness distribution of the reconstructed outputs need not match that of the input RGB frames.
To respect the task definitions and ensure fair comparison with brightness-preserving deblurring or interpolation baselines, all methods follow the same task-matched protocol: training and evaluation without exposure augmentation for deblurring and interpolation, and with exposure augmentation for HDR fusion. Comparisons of unified reconstruction using the same set of model weights across tasks are presented in Sec.~\ref{sec:real_world_generalization}.
Although the CVS provides substantially denser high-frame-rate differential measurements, we adopt a seven-frame reconstruction setting in the main comparison to balance temporal resolution and computational cost. Scaling to higher reconstruction frame rates is further evaluated in Sec.~\ref{sec:higher_frame_rate_scaling}.

\subsubsection{Motion Deblurring}
\label{sec:exp_deblurring}
We evaluate four motion-blur levels on the SportsSloMo-CVS test set, where each blurry RGB input is formed by integrating 5, 7, 9, or 11 consecutive sharp frames. For each exposure window, the sharp RGB frame at its temporal midpoint is used as the ground truth. The quantitative results are summarized in Table~\ref{tab:comparison_deblur}.

\textbf{Results.} As shown in Table~\ref{tab:comparison_deblur}, our method consistently outperforms IGFNet, the strongest general-purpose CVS reconstruction baseline, across all blur levels and evaluation metrics. Averaged over the four settings, our method improves PSNR and SSIM over IGFNet by 4.40~dB and 0.0456, respectively, while reducing LPIPS by 71.7\%. These results demonstrate that CoGenRec achieves higher pixel fidelity and perceptual quality than existing general-purpose CVS reconstruction methods across different degrees of motion blur. STGDNet, which is specifically designed for motion deblurring, achieves higher PSNR and SSIM than our method, whereas our method achieves better LPIPS at every blur level and on average. This comparison indicates that the task-specific deblurring network has an advantage in pixel-wise distortion metrics, whereas our method provides greater perceptual fidelity. Fig.~\ref{fig:qualitative_comparison}(a) further presents qualitative deblurring results for blurry RGB frames. Our method effectively suppresses motion trails and reduces color mixing artifacts, producing clearer and more vivid reconstruction results. These qualitative results highlight the perceptual advantages of our generative formulation over the regression-based baselines.

\begin{table}[!t]
\centering
\tablebodyfont
\setlength{\tabcolsep}{2pt}
\caption{Quantitative comparison of HDR fusion on the SportsSloMo-CVS test set. PSNR, SSIM, LPIPS, and NIQE are first averaged over the seven reconstructed frames within each sample and then across all samples, while FID is computed over the pooled set of all reconstructed frames. The best and second-best results in each column are highlighted in bold and underlined, respectively.}
\begin{tabular}{c|c c c c c|c}
\toprule
Method & PSNR$\uparrow$ & SSIM$\uparrow$ & LPIPS$\downarrow$ & NIQE$\downarrow$ & FID$\downarrow$ & TR$\downarrow$ \\
\midrule
Tianmouc-UNet~\cite{Tianmouc} & \underline{24.69} & 0.8416 & 0.2340 & 6.66 & 57.21 & \underline{0.022850} \\
IGFNet~\cite{lin2026brain} & 24.57 & \underline{0.8511} & \underline{0.2178} & 6.72 & 52.21 & 0.023154 \\
CBRDM~\cite{meng2025diffusion} & 17.81 & 0.7069 & 0.2445 & \textbf{5.16} & \underline{48.27} & 0.028478 \\
CoGenRec (Ours) & \textbf{29.72} & \textbf{0.9263} & \textbf{0.0647} & \underline{5.69} & \textbf{8.70} & \textbf{0.016891} \\
\bottomrule
\end{tabular}
\label{tab:comparison_hdr}
\end{table}

\begin{table*}[!t]
\centering
\tablebodyfont
\setlength{\tabcolsep}{3.0pt}
\renewcommand{\arraystretch}{1.06}
\caption{
No-reference image quality and video temporal stability evaluation on real CVS scenes. (a) Single-frame results at AOP=0. (b) Mean image quality scores over all reconstructed frames and motion-compensated temporal residuals (TR) over adjacent pairs. ``--'' denotes a metric inapplicable to single-frame methods. ``Official'' denotes the officially released checkpoint, whereas ``Retrained'' denotes a checkpoint retrained on our large-scale mixed dataset. The best and second-best results in each panel are highlighted in bold and underlined, respectively.
}
\label{tab:real_world_nr_iqa}
\begin{tabular}{c|c|c c c c c|c}
\toprule
Method & Training
& NIQE$\downarrow$
& MUSIQ$\uparrow$
& MANIQA$\uparrow$
& CLIP-IQA+$\uparrow$
& TOPIQ-NR$\uparrow$
& TR$\downarrow$ \\
\midrule

\multicolumn{8}{c}{\textit{(a) Single Frame at AOP=0}} \\
\midrule

\multirow{3}{*}{Tianmouc-UNet~\cite{Tianmouc}}
& Official Checkpoint                & 19.01 & 20.03 & 0.1279 & 0.1759 & 0.1602 & -- \\
& Retrained w/o Exp. Aug. & 7.71  & 30.78 & 0.1737 & 0.2754 & 0.2248 & -- \\
& Retrained w/ Exp. Aug.  & 7.89  & 30.79 & 0.1828 & 0.2786 & 0.2282 & -- \\
\midrule

\multirow{3}{*}{IGFNet~\cite{lin2026brain}}
& Official Checkpoint                & 16.77 & 20.28 & 0.1295 & 0.1737 & 0.1619 & -- \\
& Retrained w/o Exp. Aug. & 8.00  & 32.55 & 0.1810 & 0.3031 & 0.2275 & -- \\
& Retrained w/ Exp. Aug.  & 8.27  & 33.97 & 0.1923 & 0.3137 & 0.2422 & -- \\
\midrule

\multirow{3}{*}{CBRDM~\cite{meng2025diffusion}}
& Official Checkpoint                & 8.67 & 29.85 & 0.1915 & 0.2936 & 0.2225 & -- \\
& Retrained w/o Exp. Aug. & 6.44 & 37.40 & 0.2060 & 0.3427 & 0.2883 & -- \\
& Retrained w/ Exp. Aug.  & 6.51 & \underline{39.55} & \textbf{0.2225} & \textbf{0.3672} & \underline{0.2994} & -- \\
\midrule

\multirow{3}{*}{STGDNet~\cite{meng2026deblur}}
& Official Checkpoint                & 19.96 & 22.60 & 0.1315 & 0.2183 & 0.1686 & -- \\
& Retrained w/o Exp. Aug. & \underline{6.02} & 31.84 & 0.1907 & 0.3122 & 0.2887 & -- \\
& Retrained w/ Exp. Aug.  & \textbf{5.32} & 33.64 & 0.2046 & 0.3433 & 0.2936 & -- \\
\midrule

\multirow{2}{*}{CoGenRec (Ours)}
& w/o Exp. Aug. & 6.27 & 37.91 & 0.2018 & 0.3418 & 0.2881 & -- \\
& w/ Exp. Aug.  & 6.40 & \textbf{40.01} & \underline{0.2197} & \underline{0.3664} & \textbf{0.3037} & -- \\
\midrule

\multicolumn{8}{c}{\textit{(b) All Reconstructed Frames}} \\
\midrule

\multirow{3}{*}{Tianmouc-UNet~\cite{Tianmouc}}
& Official Checkpoint                & 18.32 & 20.20 & 0.1314 & 0.1754 & 0.1622 & 0.0165 \\
& Retrained w/o Exp. Aug. & 7.71  & 30.98 & 0.1759 & 0.2745 & 0.2275 & 0.0211 \\
& Retrained w/ Exp. Aug.  & 7.88  & 31.06 & 0.1839 & 0.2776 & 0.2299 & 0.0216 \\
\midrule

\multirow{3}{*}{IGFNet~\cite{lin2026brain}}
& Official Checkpoint                & 16.63 & 20.38 & 0.1307 & 0.1738 & 0.1643 & \textbf{0.0162} \\
& Retrained w/o Exp. Aug. & 7.99  & 32.83 & 0.1838 & 0.3040 & 0.2317 & 0.0200 \\
& Retrained w/ Exp. Aug.  & 8.22  & 34.14 & 0.1944 & 0.3117 & 0.2450 & 0.0206 \\
\midrule

\multirow{3}{*}{CBRDM~\cite{meng2025diffusion}}
& Official Checkpoint                & 8.62 & 29.37 & 0.1876 & 0.2839 & 0.2166 & 0.0190 \\
& Retrained w/o Exp. Aug. & \underline{6.47} & 38.05 & 0.2121 & 0.3395 & 0.2940 & 0.0375 \\
& Retrained w/ Exp. Aug.  & 6.61 & \underline{39.69} & \underline{0.2238} & \underline{0.3592} & \underline{0.3022} & 0.0295 \\
\midrule

\multirow{2}{*}{CoGenRec (Ours)}
& w/o Exp. Aug. & \textbf{6.44} & 38.86 & 0.2080 & 0.3479 & 0.2909 & 0.0168 \\
& w/ Exp. Aug.  & 6.56 & \textbf{40.86} & \textbf{0.2249} & \textbf{0.3691} & \textbf{0.3073} & \underline{0.0163} \\
\bottomrule
\end{tabular}
\end{table*}

\subsubsection{Frame Interpolation}
\label{sec:exp_interpolation}
For frame interpolation, we compare our method with Tianmouc-UNet~\cite{Tianmouc}, IGFNet~\cite{lin2026brain}, and CBRDM~\cite{meng2025diffusion}, evaluating image quality at the temporal midpoint and over all seven reconstructed frames on the SportsSloMo-CVS test set. Beyond single-image quality, we measure video continuity using a motion-compensated temporal residual (TR, lower is better): RAFT-Large~\cite{teed2020raft} estimates bidirectional flow between each method's adjacent outputs, and the mean absolute error in $[0,1]$ RGB is computed after bilinear warping. Table~\ref{tab:comparison_interpolation} presents the results.

\textbf{Results.} As shown in Table~\ref{tab:comparison_interpolation}, our method achieves the best PSNR, SSIM, LPIPS, and FID under both the middle-frame and all-frame settings. Across all reconstructed frames, it improves PSNR over IGFNet by 3.89~dB while reducing LPIPS and FID by 69.2\% and 76.0\%, respectively, demonstrating superior reconstruction fidelity, perceptual similarity, and distributional consistency. CoGenRec also reduces TR by 11.5\% relative to the strongest baseline, IGFNet, indicating smaller motion-compensated inter-frame variations.
Fig.~\ref{fig:qualitative_comparison}(b) shows interpolation results at the temporal midpoint between two adjacent blurry RGB endpoint frames, RGB0 and RGB1. Compared with the retrained baselines, our method more accurately recovers the positions and appearances of fast-moving objects at the intermediate time while substantially reducing ghosting and color misalignment.

\begin{figure*}[!t]
    \centering
    \includegraphics[width=0.99\textwidth]{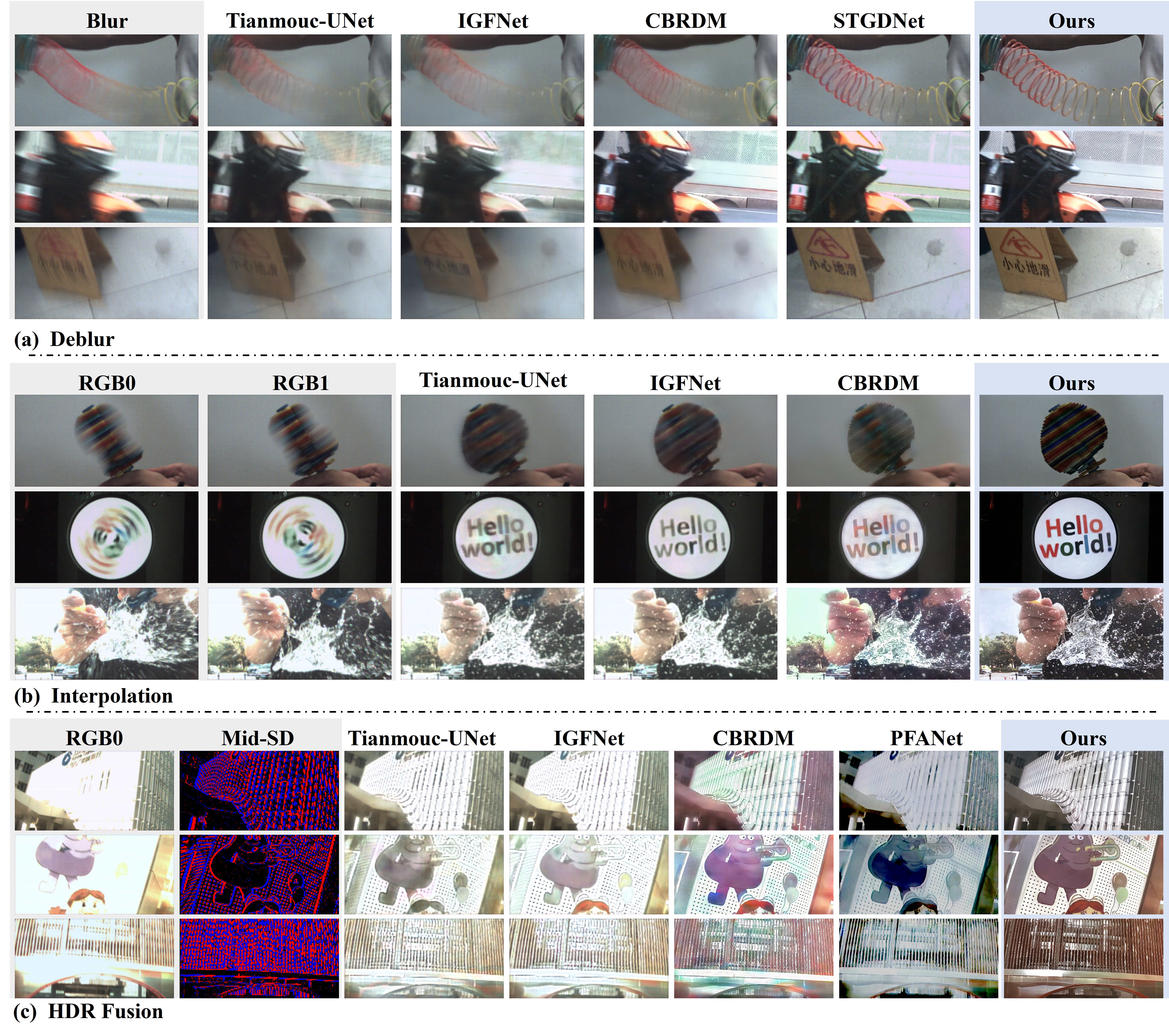}
    \caption{Qualitative comparisons on real CVS captures. All baselines use their officially released checkpoints, whereas our method uses the checkpoint trained with exposure augmentation. Each row corresponds to one real scene. (a) RGB-frame deblurring. (b) Intermediate-frame interpolation between adjacent RGB endpoints. (c) HDR fusion at the intermediate time, where the long-exposure RGB provides color and shadow information and the short-exposure Mid-SD retains structural cues in saturated highlight regions.}
    \label{fig:real_world_comparison}
\end{figure*}

\subsubsection{HDR Fusion}
\label{sec:exp_hdr_fusion}
We further evaluate the ability of each method to exploit exposure complementarity between the two sensing pathways on the SportsSloMo-CVS test set. Following the interpolation setting, each method reconstructs seven uniformly spaced target frames.
Table~\ref{tab:comparison_hdr} compares our method with Tianmouc-UNet~\cite{Tianmouc}, IGFNet~\cite{lin2026brain}, and CBRDM~\cite{meng2025diffusion}.

\textbf{Results.} As shown in Table~\ref{tab:comparison_hdr}, our method achieves the best PSNR, SSIM, LPIPS, and FID on the exposure-augmented test set. Compared with the strongest baseline for each metric, it improves PSNR by 5.04~dB and SSIM by 0.0753, while reducing LPIPS and FID by 70.3\% and 82.0\%, respectively. CoGenRec also reduces TR by 26.1\% relative to the strongest baseline, Tianmouc-UNet, indicating smaller motion-compensated inter-frame variations. These results demonstrate that our method effectively exploits the short-exposure differential measurements to compensate for highlight structures lost in the saturated RGB endpoints, while preserving substantially higher pixel fidelity, perceptual similarity, and distributional consistency. Although CBRDM obtains the lowest NIQE, our method ranks second on this metric, whereas CBRDM performs substantially worse in PSNR, SSIM, and LPIPS, indicating that a no-reference naturalness metric alone does not characterize faithful recovery of the sharp ground truth before exposure augmentation. Overall, our method provides a better balance between HDR reconstruction accuracy and perceptual quality. 
Fig.~\ref{fig:qualitative_comparison}(c) further presents HDR fusion results at the temporal position corresponding to Mid-SD between two adjacent RGB frames. Although the highlight regions in the RGB observation are severely saturated, the short-exposure differential pathway retains complementary structural cues in Mid-SD. The regression-based Tianmouc-UNet and IGFNet can also exploit these cues to recover textures in saturated regions, but their reconstruction results show color loss and muted tones, consistent with the tendency of regression-based reconstruction to favor averaged solutions. The generative baseline CBRDM produces richer colors but introduces color mixing artifacts. In comparison, our method recovers missing scene structures with vivid and accurate colors consistent with the RGB context while avoiding color mixing. These results highlight the effectiveness of combining CVS sensor-grounded constraints with the generative capacity of our model to achieve faithful structural and color recovery.

\subsection{Task-unified Real-hardware Generalization}
\label{sec:real_world_generalization}

We further evaluate all methods on 129 real CVS scenes containing complex motion and extreme illumination. Since these sequences do not provide sharp RGB ground truth, we conduct quantitative evaluation using no-reference image quality and video temporal stability metrics.
We evaluate the official checkpoints of competing methods, along with the checkpoints of all methods, including ours, trained with and without exposure augmentation in Sec.~\ref{sec:benchmark_comparisons}.

We use five image quality metrics: NIQE~\cite{mittal2013niqe}, MUSIQ~\cite{ke2021musiq}, MANIQA~\cite{yang2022maniqa}, CLIP-IQA+~\cite{wang2023clipiqa}, and TOPIQ-NR~\cite{chen2024topiq}. Lower values are preferred for NIQE, whereas higher values are preferred for the remaining four metrics. The single-frame evaluation uses AOP=0, which corresponds to the exposure center of the first blurry RGB frame for both the multi-frame methods and STGDNet. For the all-frame evaluation, all reconstructed frames are scored independently before averaging; STGDNet is excluded. For real-world temporal residuals (TR), we follow the TR computation in Sec.~\ref{sec:exp_interpolation}.

As shown in Table~\ref{tab:real_world_nr_iqa}, exposure augmentation improves all four learned perceptual metrics for our method and all retrained competing networks. Retraining competing networks on our large-scale mixed dataset also improves NIQE and learned perceptual metrics over their official checkpoints.

\begin{table*}[!t]
\centering
\tablebodyfont
\setlength{\tabcolsep}{1.5pt}
\renewcommand{\arraystretch}{1.10}
\caption{Component ablation on the SportsSloMo-CVS and Ego4D-CVS datasets. All metrics are aggregated over seven reconstructed frames.}
\label{tab:ablation_study}

\resizebox{\textwidth}{!}{
\begin{tabular}{cccc|cccccc|cccccc}
\toprule
\multicolumn{4}{c|}{Components}
& \multicolumn{6}{c|}{w/o Exposure Aug.}
& \multicolumn{6}{c}{w/ Exposure Aug.} \\
\cmidrule(lr){1-4} \cmidrule(lr){5-10} \cmidrule(lr){11-16}

\makecell{Guidance\\Extractor}
& \makecell{Spatial Diff.\\Backbone}
& \makecell{Video Temporal\\Layers}
& \makecell{Detail-Preserving\\VAE Decoder}
& PSNR$\uparrow$ & SSIM$\uparrow$ & LPIPS$\downarrow$ & NIQE$\downarrow$ & FID$\downarrow$ & TR$\downarrow$
& PSNR$\uparrow$ & SSIM$\uparrow$ & LPIPS$\downarrow$ & NIQE$\downarrow$ & FID$\downarrow$ & TR$\downarrow$ \\
\midrule

\multicolumn{16}{c}{\textit{SportsSloMo-CVS}} \\
\midrule

GT
& -- & -- & --
& $\infty$ & 1.000 & 0.000 & 5.21 & -- & 0.0131
& $\infty$ & 1.000 & 0.000 & 5.21 & -- & 0.0131 \\

$\checkmark$ & $\times$      & $\times$      & $\times$      & 30.56 & 0.9197 & 0.0653 & 5.50 & 9.99  & 0.0186 &  28.50 & 0.9060 & 0.0936 & 5.48 & 18.41 & 0.0202 \\
$\checkmark$ & $\checkmark$  & $\times$      & $\times$      & 25.78 & 0.8211 & 0.1256 & 6.31 & 14.39  & 0.0330 &  23.53 & 0.7969 & 0.1645 & 6.10 & 17.40 & 0.0511 \\
$\checkmark$ & $\checkmark$  & $\checkmark$  & $\times$      & 26.98 & 0.8407 & 0.0753 & \textbf{4.78} & 7.56  & 0.0209 &  24.87 & 0.8244 & 0.0961 & \textbf{4.75} & 9.70 & 0.0211 \\
$\checkmark$ & $\checkmark$  & $\checkmark$  & $\checkmark$  & \textbf{31.66} & \textbf{0.9355} & \textbf{0.0475} & 5.75 & \textbf{6.50}  & \textbf{0.0162} &  \textbf{29.72} & \textbf{0.9263} & \textbf{0.0647} & 5.69 & \textbf{8.70} & \textbf{0.0171} \\

\midrule
\multicolumn{16}{c}{\textit{Ego4D-CVS}} \\
\midrule

GT
& -- & -- & --
& $\infty$ & 1.000 & 0.000 & 4.83 & -- & 0.0126
& $\infty$ & 1.000 & 0.000 & 4.83 & -- & 0.0126 \\

$\checkmark$ & $\times$      & $\times$      & $\times$      & 31.60 & 0.8783 & 0.0742 & 5.58 & 12.44  & 0.0165 &  29.90 & 0.8664 & 0.0959 & 5.55 & 17.91 & 0.0175 \\
$\checkmark$ & $\checkmark$  & $\times$      & $\times$      & 25.68 & 0.7465 & 0.1647 & 5.14 & 16.28  & 0.0350 &  24.89 & 0.7563 & 0.1712 & 5.33 & 19.59 & 0.0392 \\
$\checkmark$ & $\checkmark$  & $\checkmark$  & $\times$      & 27.79 & 0.7749 & 0.0969 & \textbf{4.78} & 10.36  & 0.0223 &  26.22 & 0.7589 & 0.1231 & \textbf{4.75} & \textbf{13.21} & 0.0225 \\
$\checkmark$ & $\checkmark$  & $\checkmark$  & $\checkmark$  & \textbf{32.19} & \textbf{0.8901} & \textbf{0.0612} & 5.79 & \textbf{10.02}  & \textbf{0.0152} &  \textbf{30.39} & \textbf{0.8790} & \textbf{0.0781} & 5.66 & 13.79 & \textbf{0.0159} \\

\bottomrule
\end{tabular}
}
\end{table*}

Our method outperforms official competing checkpoints on all five image quality metrics in both single-frame and all-frame settings. Against retrained models, our exposure-augmented model still achieves the best scores on all four learned metrics across all reconstructed frames and on MUSIQ and TOPIQ-NR for single frames. For video temporal consistency, this model achieves a TR of 0.0163, slightly higher than official IGFNet's 0.0162, but with substantially better perceptual image quality across all reconstructed frames.

Figure~\ref{fig:real_world_comparison}(a) presents real RGB-frame deblurring results. STGDNet and our method better preserve scene colors and moving-object boundaries while suppressing motion trails in the RGB inputs; our method further exploits the exposure complementarity of the two CVS pathways to reduce highlight saturation. Figure~\ref{fig:real_world_comparison}(b) shows intermediate-frame results between adjacent RGB frames. In extreme-motion scenes involving rapidly rotating disks and blocks, the competing methods produce color mixing, desaturation, or local chromatic inconsistency, whereas our method preserves both sharp motion structures and coherent colors. In the water-droplet scene, our method further uses the short-exposure difference signals to recover structures in regions saturated in the RGB endpoints.
Figure~\ref{fig:real_world_comparison}(c) further shows the dynamic-range complementarity of the two CVS pathways. The long-exposure RGB observation preserves scene colors and shadow information, while the short-exposure Mid-SD retains fine structures in strongly illuminated regions. Tianmouc-UNet and IGFNet produce washed-out highlights, and CBRDM exhibits noticeable color mixing and bleeding, whereas our method recovers clearer highlight structures and more complete colors.

Overall, these real-world results demonstrate that our large-scale paired supervision improves practical reconstruction performance across diverse models, including the competing baselines, while exposure augmentation further enhances perceptual quality under challenging illumination. Building on these benefits, our generative framework effectively combines CVS observation constraints with generative capacity to recover sharp scene structures and produce high-quality reconstructions with vivid, observation-consistent colors.

\subsection{Ablation Study}

\textbf{Ablation Setup.} We progressively construct CoGenRec to examine the key designs for scene reconstruction under extreme dynamic conditions: sensor observation constraints, generative appearance recovery, temporal consistency, and detail preservation. Under these conditions, fast motion, temporal undersampling, and exposure saturation degrade or obscure visual information, requiring reconstruction to exploit sensor measurements to constrain scene content while drawing on generative priors to recover appearance. We begin with deterministic reconstruction using the Multi-Scale Spatio-Temporal Guidance Extractor and its Stage-1 auxiliary decoder. We then introduce the Spatial Diffusion Backbone, comprising a pre-trained image UNet and a multi-scale ControlNet, followed by the Video Temporal Layers. Finally, we incorporate the Detail-Preserving VAE Decoder.

For a fair comparison, all ablation models are trained on the complete mixed training set described in Sec.~\ref{sec:datasets}, both with and without exposure augmentation. Table~\ref{tab:ablation_study} reports the results on SportsSloMo-CVS and Ego4D-CVS. PSNR, SSIM, LPIPS, and NIQE are first averaged over the seven reconstructed frames of each sample and then across all samples, whereas FID is computed over all reconstructed frames pooled across the test set. TR follows the computation in Sec.~\ref{sec:exp_interpolation}.

\begin{figure*}[!t]
    \centering
    \includegraphics[width=0.8\textwidth]{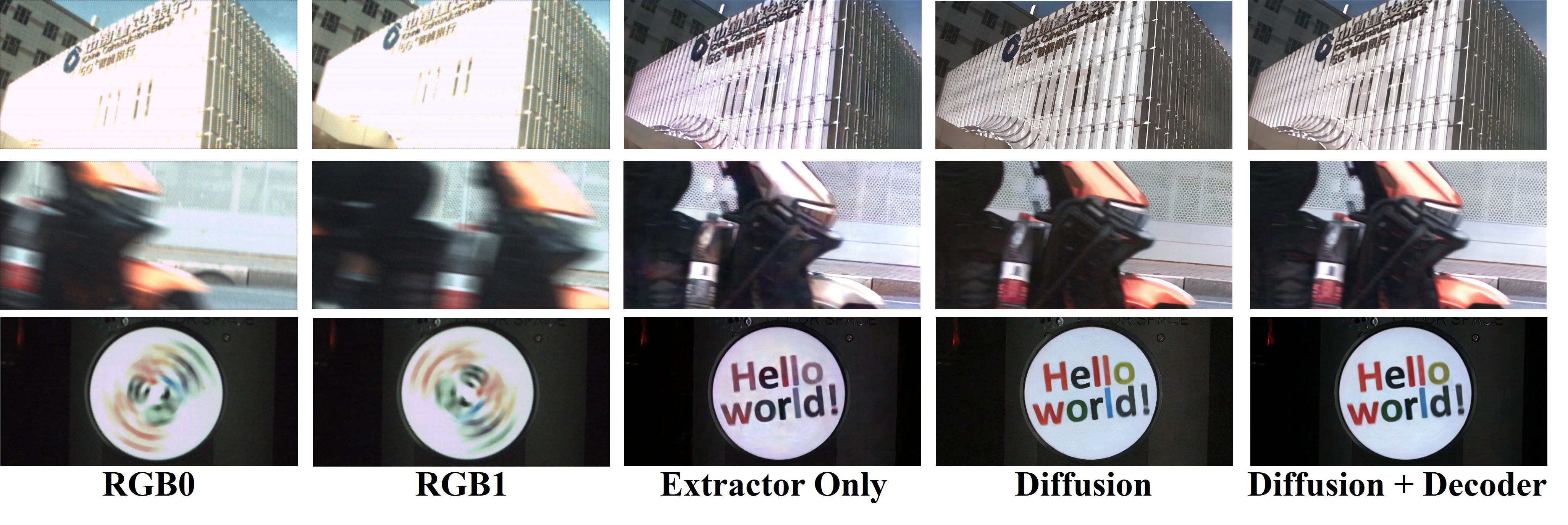}
    \caption{Qualitative ablation in HDR scenes and under severe motion blur. RGB0 and RGB1 are the input frames. \textbf{Extractor Only} reconstructs images directly through the Multi-Scale Spatio-Temporal Guidance Extractor and its Stage-1 auxiliary decoder. \textbf{Diffusion} adds the Stable Diffusion spatial backbone and Video Temporal Layers, conditioned through the multi-scale ControlNet and decoded with the original VAE decoder. \textbf{Diffusion + Decoder} further adds the Detail-Preserving VAE Decoder. With the extractor alone, the building highlights look slightly red, while regions affected by severe motion blur exhibit faded and mixed colors. Diffusion produces more vivid and distinct colors; the detail-preserving decoder further refines local structures, including text strokes and edges, while retaining these color improvements.}
    \label{fig:ablation_visual}
\end{figure*}

\textbf{Observation-Grounded Reconstruction.} The extractor-only model already achieves strong pixel-level results, reaching PSNR values of 30.56 and 31.60~dB on SportsSloMo-CVS and Ego4D-CVS, respectively, without exposure augmentation. This demonstrates that translating complementary sensor measurements into dense structural and motion representations provides an effective foundation for dynamic scene reconstruction. However, Fig.~\ref{fig:ablation_visual} shows that this deterministic variant still produces reddish building highlights in the HDR scene and faded or mixed colors in regions affected by severe motion blur. Strong pixel-level accuracy therefore does not imply complete appearance recovery in degraded regions. Observation-derived guidance provides scene-specific structural and motion evidence, but reconstruction ambiguities caused by severe motion blur, highlight saturation, and temporal undersampling remain difficult to resolve through deterministic regression alone, particularly where reliable color and texture information has been lost. This illustrates the limitations of purely regression-based reconstruction under highly degraded observations.

\begin{figure*}[!t]
    \centering
    \includegraphics[width=0.99\textwidth]{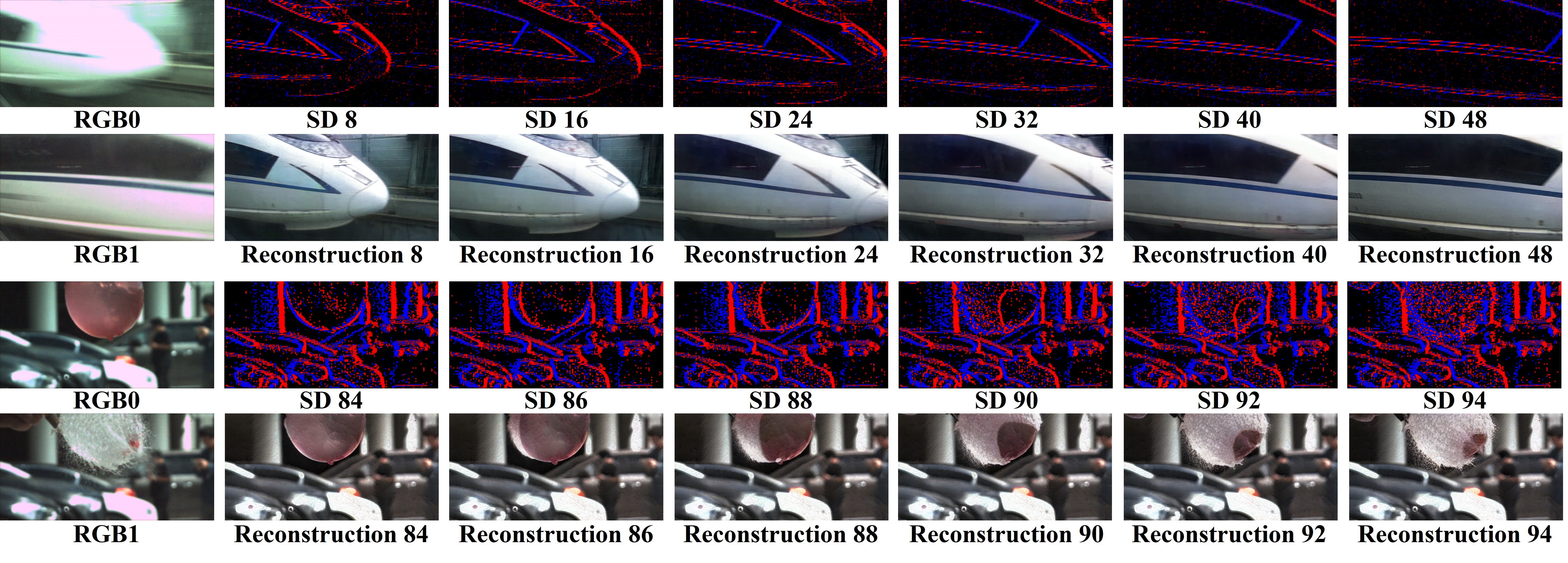}
    \caption{Reconstruction results from real CVS captures under different frame-rate modes. The first two rows show two high-speed trains passing each other, for which the spatio-temporal difference pathway operates at 1515~FPS and acquires 50 differential measurements between RGB0 and RGB1; reconstructions at selected time instants are shown. The last two rows show a punctured water balloon, for which the pathway operates at 3030~FPS and acquires 100 differential measurements between RGB0 and RGB1; the selected key frames capture the progressive retraction of the balloon membrane after rupture.}
    \label{fig:higher_frame_rate_scaling}
\end{figure*}

\textbf{Generative Priors for Appearance Recovery.} Compared with the extractor-only model, the variant containing both the Spatial Diffusion Backbone and Video Temporal Layers, but retaining the original VAE decoder, achieves lower NIQE and FID on both datasets under standard and exposure-augmented settings, indicating improved image naturalness and distributional similarity. For example, on exposure-augmented SportsSloMo-CVS, FID decreases from 18.41 to 9.70. Although PSNR decreases from 28.50 to 24.87~dB at this stage, subsequent observation-guided decoding raises it to 29.72~dB in the complete model, surpassing the extractor-only baseline. The qualitative results further show that, even before introducing the detail-preserving decoder, the video diffusion backbone reduces reddish highlights, desaturation, and color mixing, recovering more vivid and distinct colors. These results validate the role of generative priors in appearance recovery within degraded regions, while showing that improved generative quality must be coupled with fine-grained observation-guided reconstruction to translate into greater scene fidelity.

\textbf{Temporal Consistency in Generative Reconstruction.} Introducing the Spatial Diffusion Backbone alone does not consistently improve perceptual metrics and increases TR relative to the extractor-only model in every setting. Adding the Video Temporal Layers reduces TR by 36.1--58.7\% relative to the spatial-only variant, while improving LPIPS and FID in all settings. These results validate the effectiveness of explicitly modeling cross-frame dependencies during generation. Temporal propagation in the guidance extractor organizes sensor-measured motion into temporally aligned conditions, while the temporal layers in the diffusion backbone further coordinate appearance generation across frames. Together, they facilitate coherent video reconstruction under extreme dynamic conditions.

\textbf{Observation-Guided Detail Recovery.} The complete model with the Detail-Preserving VAE Decoder achieves the best PSNR, SSIM, LPIPS, and TR among all reconstruction variants on both datasets under standard and exposure-augmented settings. On exposure-augmented SportsSloMo-CVS, adding this decoder improves PSNR by 4.85~dB and reduces LPIPS by 32.7\% relative to the variant without it. Through skip connections, the decoder incorporates multi-scale guidance features that bypass VAE latent-space compression, allowing sensor observations to contribute directly to fine-structure recovery in addition to constraining generation. As shown in Fig.~\ref{fig:ablation_visual}, the complete model retains the vivid colors recovered by the diffusion backbone while further refining text strokes and edges on the rotating disk. These results validate the effectiveness of reintroducing observation-derived features during final decoding, making generative appearance recovery and fine-grained structural reconstruction complementary. Taken together, the quantitative and qualitative results demonstrate a better balance among reconstruction accuracy, perceptual quality, and temporal stability in the complete framework.

Overall, these ablations validate the complementary roles of CoGenRec's designs for scene reconstruction under extreme dynamic conditions: sensor ground structure and motion, generative priors help recover degraded or missing appearance information, temporal modeling coordinates generation across frames, and observation-guided decoding restores fine details. Their integration enables improved colors and perceptual quality while maintaining strong scene fidelity and temporal consistency.

\subsection{Real-World Higher Frame Rate Performance}
\label{sec:higher_frame_rate_scaling}

The spatio-temporal difference pathway of the current CVS supports multiple frame-rate and quantization-precision configurations under a configurable rate--precision trade-off, including 757~FPS at $\pm7$-bit, 1515~FPS at $\pm7$-bit, and 3030~FPS at $\pm4$-bit. To evaluate how well the model scales to higher reconstruction frame rates, we further test it on scenes with extremely fast motion captured using real CVS hardware. 
For these experiments, we use a variable-length reconstruction model fine-tuned from the seven-frame model (in Sec.~\ref{sec:benchmark_comparisons}) to accommodate these input modes. As shown in Fig.~\ref{fig:higher_frame_rate_scaling}, CoGenRec reconstructs two high-speed trains passing each other under the 1515~FPS setting and recovers the instant of water-balloon rupture and the rapid retraction of its membrane under the 3030~FPS setting. These results demonstrate that our method can accommodate different CVS rate--precision configurations and recover transient motion details that are difficult to observe at conventional video frame rates.


\section{Conclusion}
We have presented CoGenRec, a complementarity-driven generative reconstruction framework that unifies video deblurring, frame interpolation, and HDR fusion for complementary vision sensors. CoGenRec exploits three forms of CVS complementarity, complementary spatio-temporal sampling, the decoupling of high-SNR imaging from fast-motion sensing, and exposure complementarity, to integrate multiscale sensor-grounded guidance with pretrained video diffusion priors and detail-preserving decoding. This design uses physical measurements to constrain scene appearance, structure and temporal evolution while leveraging learned video priors to recover texture details that are weakly observed or lost during acquisition.
Trained on a hybrid dataset of approximately one million hardware-converted and noise-aware simulated clips, CoGenRec consistently outperforms existing general-purpose CVS reconstruction methods across all three tasks and achieves strong perceptual quality on 129 unseen real-world CVS scenes. These results demonstrate the potential of combining complementary physical measurements with pretrained generative priors to reconstruct sharp, temporally dense, and visually detailed video under challenging motion and illumination conditions. Beyond CVS reconstruction, this approach may inform computational imaging systems for applications such as autonomous perception, high-speed scientific imaging, and robotic vision, where reliable visual information must be recovered under demanding motion and lighting conditions.

\ifCLASSOPTIONcompsoc
  \section*{Acknowledgments}
\else
  \section*{Acknowledgment}
\fi

This work was supported by the National Key Research and Development Program of China (No. 2025YFG0100200) and the Tsinghua University Initiative Scientific Research Program 20257020014.

\ifCLASSOPTIONcaptionsoff
  \newpage
\fi


\newpage

\bibliographystyle{IEEEtran}
\bibliography{refs}
%



%





\end{document}